%% file: main.tex
\documentclass[11pt]{article}

\usepackage[final]{acl}

\usepackage{times}
\usepackage{latexsym}

\usepackage[T1]{fontenc}

\usepackage[utf8]{inputenc}

\usepackage{microtype}

\usepackage{inconsolata}

\usepackage{graphicx}
\usepackage[colorinlistoftodos]{todonotes}

\title{The Geometry of Ignorance: Detecting Suppressed Knowledge in LLMs through Activation Fragmentation}

\author{
  \textbf{Adar Peleg\textsuperscript{*}},
  \textbf{Dvir Alsheich\textsuperscript{*}},
  \textbf{Tomer Ashuach},
  \textbf{Avi Mendelson}
\\
  \textsuperscript{1}Technion - Israel Institute of Technology
\\
  \textsuperscript{*}Equal contribution
\\
  \small{
    \textbf{Correspondence:} \texttt{\{a.p, dviralsheich, tomerashuach\}@campus.technion.ac.il}, \texttt{mendlson@technion.ac.il}
  }
}

\usepackage{amsmath}
\usepackage{amssymb}
\usepackage{amsthm}
\newtheorem{theorem}{Theorem}
\newtheorem*{theorem*}{Theorem}
\newtheorem{proposition}{Proposition}
\newtheorem{corollary}{Corollary}
\newtheorem{lemma}{Lemma}
\newtheorem*{remark}{Remark}
\usepackage{multirow}
\usepackage{makecell}

\begin{document}
\maketitle
\input{Sections/00-abstract}
\input{Sections/01-introduction}
\input{Sections/02-background}
\input{Sections/03-method}

\input{Sections/04-expirements}
\input{Sections/05-discussion}
\input{Sections/06-related}
\input{Sections/07-Limitations}

\bibliography{referencesV2}

\input{Sections/08-appendix}

\end{document}

%% file: Sections/00-abstract.tex
\begin{abstract}
Evaluating whether a large language model (LLM) genuinely knows a concept remains a fundamental problem, since behavioral metrics can be fooled by suppression: a model that refuses to answer may still harbor recoverable knowledge.
The framework measures the geometric coherence of \textit{anon-vectors}, the differences in hidden states between a concept-specific query and its anonymized paraphrase, and shows that genuine knowledge forms a tight, low-entropy cluster while ignorance forces geometric fragmentation.
Across ten instruction-tuned LLMs from three model families, all fragmentation metrics reliably separate known from unknown concepts, and post-unlearning, fragmentation geometry correlates with adversarial recovery across four attack types and five unlearning algorithms, though the direction of that relationship is shown to depend on attack methodology.
Furthermore, a layer-resolved analysis reveals that different adversarial mechanisms exploit this geometry at varying depths, with high attack-specific correlations that are invisible to behavioral evaluation.
Together, these results constitute a single-phase, attack-free proxy for unlearning robustness that captures aspects of parametric knowledge inaccessible to output-level metrics.

A reference implementation is available at \href{https://github.com/AdarPeleg/geometry-of-ignorance}{\includegraphics[height=1em]{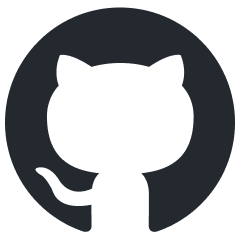}}

\end{abstract}

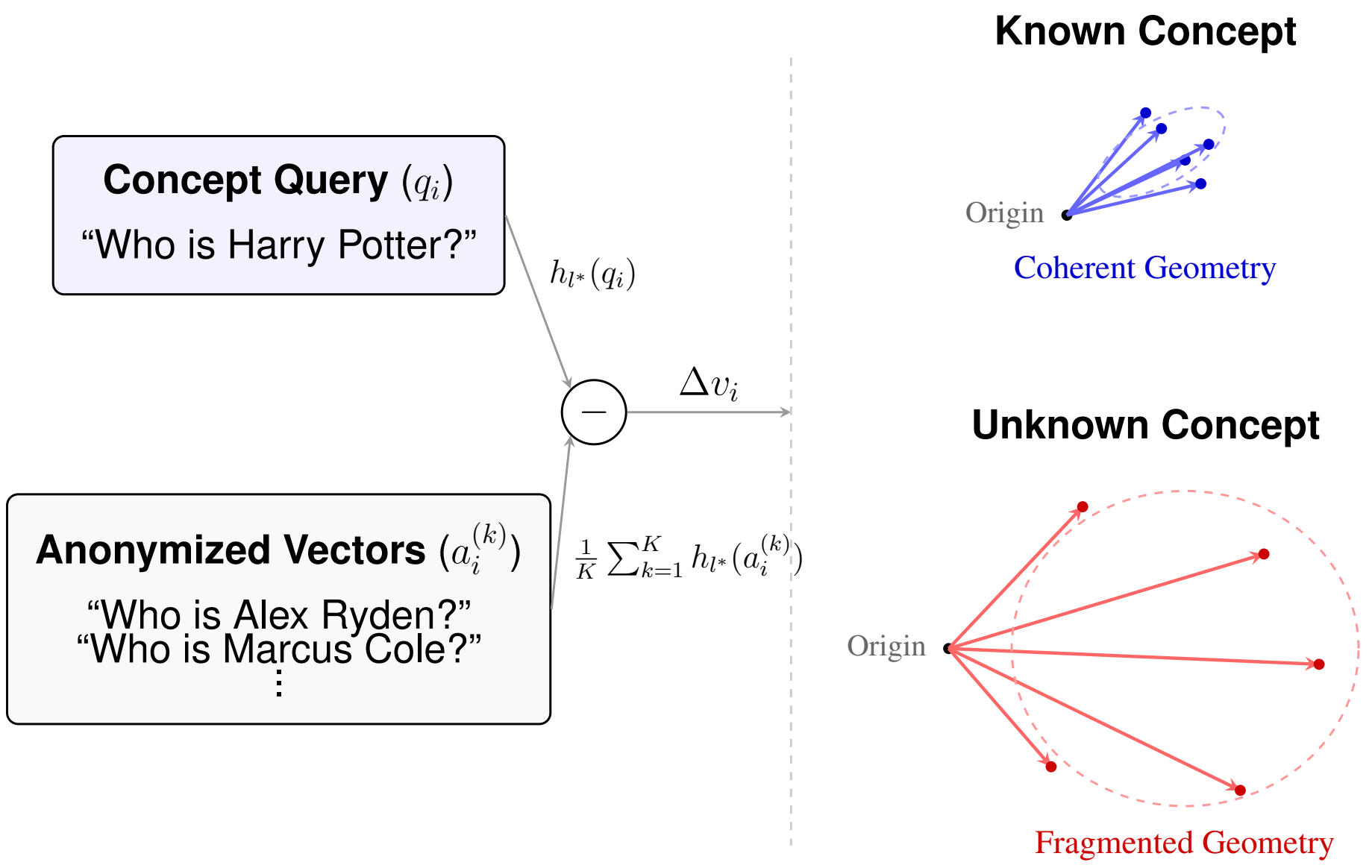
\begin{figure*}[t]
    \centering
    \resizebox{0.75\linewidth}{!}{%
    \begin{tikzpicture}[>=stealth, thick, font=\sffamily\Large]

        \node[draw, rounded corners, fill=blue!5, inner sep=10pt, align=center] (q) at (0, 2.5) {
            \textbf{Concept Query} ($q_i$)\\[1ex]
            ``Who is Harry Potter?''
        };
        
        \node[draw, rounded corners, fill=gray!5, inner sep=10pt, align=center] (a) at (0, -2.5) {
            \textbf{Anonymized Vectors} ($a_i^{(k)}$)\\[1ex]
            ``Who is Alex Ryden?''\\[-0.5ex]
            ``Who is Marcus Cole?''\\[-0.5ex]
            $\vdots$
        };
        
        \node[circle, draw, inner sep=3pt] (minus) at (4, 0) {$-$};
        
        \draw[->, color=gray!80] (q.east) -- (minus.north west) 
            node[midway, above right, text=black, font=\large] {$h_{l^*}(q_i)$};
            
        \draw[->, color=gray!80] (a.east) -- (minus.south west) 
            node[midway, below right, text=black, font=\large] {$\frac{1}{K}\sum_{k=1}^K h_{l^*}(a_i^{(k)})$};
        
        \node (dv) at (6.5, 0) {};
        \draw[->, color=gray!80] (minus.east) -- (dv.center) 
            node[midway, above, text=black] {$\Delta v_i$};
        
        \draw[dashed, color=gray!40] (6.5, 4.5) -- (6.5, -5.5);

        \node[align=center] at (11, 4.8) {\textbf{Known Concept}};
        
        \coordinate (origK) at (10, 2.5);
        \fill (origK) circle (2pt) node[left=4pt, font=\normalsize, color=gray!80!black] {Origin};
        
        \draw[->, color=blue!60, very thick] (origK) -- (11.5, 3.2); \fill[blue!80!black] (11.5, 3.2) circle (2pt);
        \draw[->, color=blue!60, very thick] (origK) -- (11.2, 3.6); \fill[blue!80!black] (11.2, 3.6) circle (2pt);
        \draw[->, color=blue!60, very thick] (origK) -- (11.7, 2.9); \fill[blue!80!black] (11.7, 2.9) circle (2pt);
        \draw[->, color=blue!60, very thick] (origK) -- (11.0, 3.8); \fill[blue!80!black] (11.0, 3.8) circle (2pt);
        \draw[->, color=blue!60, very thick] (origK) -- (11.8, 3.4); \fill[blue!80!black] (11.8, 3.4) circle (2pt);
        
        \draw[blue!40, dashed, rotate around={30:(11.2, 3.3)}] (11.2, 3.3) ellipse (0.9 and 0.4);
        \node[blue!80!black, font=\large] at (11, 1.8) {Coherent Geometry};
        
        \node[align=center] at (11, -0.2) {\textbf{Unknown Concept}};
        
        \coordinate (origU) at (8.5, -3.0);
        \fill (origU) circle (2pt) node[left=4pt, font=\normalsize, color=gray!80!black] {Origin};
        
        \draw[->, color=red!60, very thick] (origU) -- (12.5, -1.8); \fill[red!80!black] (12.5, -1.8) circle (2pt);
        \draw[->, color=red!60, very thick] (origU) -- (10.2, -1.2); \fill[red!80!black] (10.2, -1.2) circle (2pt);
        \draw[->, color=red!60, very thick] (origU) -- (12.2, -4.8); \fill[red!80!black] (12.2, -4.8) circle (2pt);
        \draw[->, color=red!60, very thick] (origU) -- (9.8, -4.5);  \fill[red!80!black] (9.8, -4.5)  circle (2pt);
        \draw[->, color=red!60, very thick] (origU) -- (13.2, -3.2); \fill[red!80!black] (13.2, -3.2) circle (2pt);
        
        \draw[red!40, dashed] (11.5, -3) ellipse (2.2 and 2.0);
        \node[red!80!black, font=\large] at (11.5, -5.5) {Fragmented Geometry};
        
    \end{tikzpicture}%
    }
    \vspace{-2mm} 
    \caption{Activation fragmentation isolates conceptual knowledge. Subtracting the mean hidden state of a cloud of anonymized variants cancels shared linguistic syntax. Plotting the resulting anon-difference vectors ($\Delta v_i$) from the origin reveals that parametric knowledge manifests as tight, low-entropy covariance clusters of aligned vectors, whereas ignorance forces geometric fragmentation across high-variance directions.}
    \label{fig:teaser}
\end{figure*}

%% file: Sections/01-introduction.tex
\section{Introduction}
\label{sec:introduction}

The expansive pre-training of Large Language Models (LLMs) inevitably leads to the memorisation of vast amounts of factual associations \cite{petroni_language_2019, meng_locating_2023}, raising critical legal, ethical, and safety concerns regarding privacy and the generation of harmful content \cite{carlini_extracting_2021, seyitoglu_extracting_2024}. Traditionally, the boundaries of a model's parametric knowledge have been evaluated through behavioral analysis, relying on surface-level metrics such as question-answering accuracy or token perplexity \cite{lynch_eight_2024}. Behavioral evaluations treat the model as a black box and fail to capture how factual associations are internally stored and retrieved during the forward pass \cite{geva_dissecting_2023}. This gap is most consequential in the context of machine unlearning, which aims to selectively erase specific concepts without retraining \cite{li_wmdp_2024, gandikota_erasing_2025, ashuach_crisp_2026}. Recent representation-level analyses reveal that unlearning interventions often result in the mere suppression or bypassing of knowledge, rather than genuine deletion \cite{xu_unlearning_2025, lee_mechanistic_2024, schoepf_redirection_2026}. Because suppression mechanisms leave the knowledge structurally distinguishable in the parameters~\cite{huu-tien_effects_2025}, it can be reliably recovered by malicious actors through activation steering~\cite{seyitoglu_extracting_2024}. Consequently, robust evaluation must shift to fine-grained, intrinsic metrics that can detect suppressed but recoverable knowledge.

Recent mechanistic interpretability work has introduced intrinsic probes that bypass generation and measure knowledge directly within intermediate model layers \cite{gottesman_estimating_2024, hong_intrinsic_nodate}. 
However, these metrics remain entangled with prompt syntax, conflating factual knowledge with the linguistic and syntactic structure of the query. Moreover, no single-phase metric reliably predicts whether suppressed knowledge is adversarially recoverable.

To address the limitations of both behavioral evaluations and syntax-entangled intrinsic metrics, we build upon the concept of the \textit{anon-vector}~\citep{seyitoglu_extracting_2024}, defined as the directional shift between the hidden state of a concept-specific query and its identity-stripped paraphrase.
We use this to introduce \textit{activation fragmentation}, a geometric framework that isolates the precise footprint of parametric knowledge using two forward passes per query (no token generation, no iterative optimization).

We hypothesize that unknown concepts lack parametric structure: their anon-difference vectors scatter into high-variance, fragmented directions. By contrast, genuine parametric knowledge manifests as a tight, cohesive geometric signature in activation space.

\paragraph{Contributions.}
Our main contributions are as follows:
\begin{itemize}
    \item We propose \textit{activation fragmentation}, a fine-grained, representation-level metric that leverages anon-vectors to successfully isolate conceptual identity from syntax in intermediate model layers.
    \item We demonstrate empirically that fragmentation metrics distinguish known entities (e.g., Harry Potter) from completely unknown fictitious concepts (e.g., TOFU \cite{maini_tofu_2024}) with robust statistical separability across ten instruction-tuned LLMs from three distinct model families.
    \item \textit{\textbf{A}ctivation \textbf{U}niformity \textbf{S}catter \textbf{S}core (AUSS)}. We establish that post-unlearning fragmentation serves as a highly accurate, attack-free predictor of elements of unlearning robustness. (e.g. AUSS dispersion metrics predict in-context learning (ICL) recoverability).
\end{itemize}

%% file: Sections/02-background.tex
\section{Background}
\label{sec:background}

\paragraph{Knowledge in LLM activations.}
LLMs encode factual associations in their parameters, with feed-forward layers
functioning as key-value memories that promote specific concepts into the residual
stream~\cite{geva_transformer_2021, geva_transformer_2022}.
Intermediate layers carry geometrically richer knowledge representations than the
final layer~\cite{skean_layer_2025, razzhigaev_shape_2023}.
\citet{skean_layer_2025} measure representational compression layer-by-layer via
the Rényi-2 Gram-matrix entropy: for any matrix $Z\in\mathbb{R}^{M\times d}$,
\begin{equation}
\begin{array}{c}
     K_Z = \frac{ZZ^{\top}}{\operatorname{tr}(ZZ^{\top})} \\
     S_2(Z) = -\log_2 \operatorname{tr}\!\left(K_Z^{2}\right) 
\end{array}
  \label{eq:gram_entropy}
\end{equation}
Lower $S_2$ indicates that the $M$ vectors concentrate in a smaller subspace
(higher representational compression), identifying where factual content is most
geometrically concentrated in the hidden space.
Applied to the raw query hidden states $\mathbf{H} = \{h_{\ell^*}(q_i)\}$,
this gives the baseline metric \textbf{Reg~$S_2$}~\citep{skean_layer_2025}:
$\mathrm{Reg}_{S_2} = S_2(\mathbf{H})$.

\paragraph{Anon-vectors and concept-differential representations.}
A concept-specific signal can be isolated by contrasting the hidden state of a
real query with that of a semantically equivalent but identity-stripped
(anonymised) paraphrase.
\citet{seyitoglu_extracting_2024} exploit this to mount an \emph{activation
steering attack}: they compute a steering vector as the raw difference between
hidden states of a concept query and its anonymised counterpart, then inject it
into the residual stream to re-activate suppressed knowledge in an already-unlearned
model.
The ConceptVectors dataset~\cite{hong_intrinsic_nodate} likewise
demonstrates that named entities encoded in model parameters leave verifiable
parametric traces at intermediate layers, and that jailbreak attacks ride along
these very directions to bypass unlearning~\cite{hong_intrinsic_nodate}.
Our anon-difference vector $\Delta v_i$ (Eq.~\ref{eq:anon_vec}) builds directly on
the anonymisation technique of~\citet{seyitoglu_extracting_2024}, averaging over
$K$ variants to stabilise the concept-differential signal.
Neither prior work, however, treats the \emph{population geometry} of a batch of
such vectors as a diagnostic metric; that is a contribution of this paper.

%% file: Sections/03-method.tex
\section{Method}
\label{sec:method}

\subsection{Metric Definitions}
\label{sec:metrics}

\paragraph{Anon-vectors and Entropy.}
Let $\{q_i\}_{i=1}^{N}$ be $N$ queries about a target concept, and let
$h_{\ell}(\cdot)\in\mathbb{R}^d$ be the mean-pooled hidden states at layer~$\ell$.
For each $q_i$, we construct $K$ anonymised variants $\{a_i^{(k)}\}_{k=1}^{K}$.
The \textbf{anon-difference vector} at layer $\ell$ isolates conceptual identity
by subtracting the mean anonymised state:
\begin{equation}
  \Delta v_i(\ell) \;=\; h_{\ell}(q_i)
    \;-\;\frac{1}{K}\sum_{k=1}^{K} h_{\ell}\!\left(a_i^{(k)}\right)
  \label{eq:anon_vec}
\end{equation}
This requires two forward passes per query in performed experiments (in general $K{+}1$ per
query, with $K{=}1$ in experiment sections) and no token generation.
All fragmentation metrics are computed from the matrix of anon-difference vectors at a chosen layer; how that layer is selected is stated in each experimental
protocol (Section~\ref{sec:experiments}) and expended in (Appendix~\ref{sec:layer_selection}).
To measure representational rank concentration, we apply Gram-matrix entropy to these unnormalised vectors, defining our primary dispersion metric:
$\mathrm{AnonDiff}_{S_2} = S_2(\Delta\mathbf{V}(\ell))$.
Applying $S_2$ to concept-differential vectors rather than raw activations is a
novel contribution of this work.
We also evaluate two variants: $\mathrm{Reg}_{S_2} = S_2(\mathbf{H}^\ell)$ applies the same entropy to the raw hidden-state matrix (no anonymisation subtraction), serving as a baseline; and $\mathrm{NormAnonDiff}_{S_2} = S_2(\hat{\Delta}\mathbf{V}(\ell))$ applies it to the $L_2$-normalised anon-difference vectors.

\paragraph{AUSS Batch-mean metrics \textbf{(Ours)}.}
To assess population-level geometric consistency, we partition the $N$ pairs into $B=10$ disjoint batches (Appendix~\ref{appendix:batch_sensitivity}). For each batch $b$, we compute the unit-normalised mean anon-difference direction $\hat{d}_b$. We then measure geometric consistency across these $B$ representative directions using two primary novel matrices that aim to measure \textit{fragmentation}:
\begin{itemize}
    \item $\mathbf{AUSS_{L_2}}$: Mean pairwise squared $L_2$ distance between batch-mean directions.
    \item $\mathbf{AUSS_{\cos^2}}$: Complement of mean pairwise squared cosine between batch-mean directions.
    \item $\mathbf{AUSS_{Jac}}$: Jaccard distance between the sets of the $k{=}32$ largest-magnitude coordinates of each pair of batch-mean directions, the fraction of top-activated dimensions that differ between batches.
\end{itemize}
Note that all three metrics are bounded: $\mathrm{AUSS}_{L_2} \in [0,2]$ and
$\mathrm{AUSS}_{\cos^2}, \mathrm{AUSS}_{Jac} \in [0,1]$.
We expect that when a concept is known, batch-mean directions cluster tightly, driving
all three metrics toward zero; when unknown, they are expected to scatter, driving them toward
the upper bound.
This bounded range means that Cohen's $|d|$ can appear large even with
modest raw gaps, because the pooled within-group standard deviation is
bounded from above.
A comprehensive list of additionally evaluated metric variations including base pairwise similarities, Jaccard incoherence, and covariance traces is detailed in Appendix~\ref{app:additional_metrics}.

\subsection{Entropy, Rank, and the Geometry of Parametric Knowledge}
\label{sec:mathematics}

The central hypothesis of this work is that parametric knowledge leaves a geometric imprint. If a model \emph{knows} a concept, diverse queries converge onto a shared subspace, producing a tight, low-entropy cluster of anon-difference vectors. Conversely, \emph{ignorance} drives these vectors to scatter across the activation space, increasing entropy.

This connection between the Gram-matrix entropy $S_2$ and geometric dispersion is formalised by the following result (for proofs see Appendix~\ref{app:proofs}):

\begin{theorem}[High entropy forces geometric fragmentation]
\label{thm:entropy_fragmentation}
Let $A \in \mathbb{R}^{n \times d}$ be a matrix of $L_2$-normalised activation vectors and let $G = AA^\top$ be the corresponding Gram matrix\footnote{For a matrix $A$ of unit-normalised vectors, the Gram matrix $G{=}AA^\top$ collects every pairwise cosine similarity ($G_{ij}{=}\cos(a_i,a_j)$), so its eigenvalue spectrum summarises how many effective directions the vectors collectively occupy.} with effective rank $\mathrm{EffRank}(G) = \mathrm{Tr}(G)/\lambda_{\max}(G)$.
Then the mean pairwise squared Euclidean distance satisfies
\[
  D_{L_2}^2 \;\geq\; 2 - \frac{2}{\mathrm{EffRank}(G)}.
\]
\end{theorem}

Since $2^{S_2/2} \leq \mathrm{EffRank}(G) \leq 2^{S_2}$ (Proposition~\ref{prop:entropy_effrank}, Appendix~\ref{app:proofs}), high $S_2$ forces the effective rank upward, which structurally guarantees spatial dispersion. 
The complementary direction holds under a natural assumption: when all
pairwise cosine similarities are non-negative (i.e., vectors share a common
half-space). For anon-vectors this is geometrically motivated:
if a model parametrically encodes a concept, every query about that
concept shifts the hidden state \emph{toward the same concept direction},
so the differences $\Delta v_i$ collectively point away from the
anonymised baseline in a consistent orientation.
Under this assumption, a matching upper bound applies:

\begin{theorem}[Low entropy forces geometric clustering]
\label{thm:clustering}
Under the same setup as Theorem~\ref{thm:entropy_fragmentation}, if additionally
$G_{ij} \geq 0$ for all $i \neq j$, then
\[
  D_{L_2}^2 \;\leq\; \frac{2n}{n-1}\!\left(1 - 2^{-S_2(G)}\right).
\]
In particular, $D_{L_2}^2 \to 0$ as $S_2(G) \to 0$.
\end{theorem}

Together, these bounds demonstrate that subspace-compression metrics directly reflect the presence or absence of a coherent concept representation, effectively isolating knowledge from mere behavioural suppression.

\subsection{Semantic Testing}
\label{sec:semantic_testing}

\paragraph{Anon-vectors over raw activations.}
Raw activations encode both conceptual content and prompt syntax. Subtracting the anonymised baseline (denoted as$\Delta v_i$) cancels shared syntactic components, isolating the conceptual signal. This produces a substantially stronger discriminative signal than working with raw activations directly (Table~\ref{tab:rq1_disc}).

\paragraph{Semantic vs.\ generation artifacts.}
Behavioral evaluations measure accessible outputs, not latent storage. Since representational suppression mechanisms can block factual recall without deleting the underlying knowledge \citep{lee_mechanistic_2024,xu_unlearning_2025}, evaluating hidden-state geometry provides a strictly more rigorous proxy for true parametric erasure. A model that refuses to answer, whether through RLHF-driven refusal or explicit unlearning, may still harbor recoverable knowledge in its hidden states; generation-based metrics will report success while the knowledge persists, making them insufficient as a sole evaluation criteria.

\paragraph{The measurement problem: no true oracle.}
The ideal ground truth for a knowledge-detection metric would be a model trained from scratch on a corpus that provably excludes the target concept, making its ignorance logically certain. In practice, this is prohibitively expensive and inaccessible: controlled pre-training runs require substantial compute, carefully curated multi-billion-token datasets, and institutional resources. There is, therefore, no true oracle for parametric knowledge. We cannot directly observe what a model knows, only what it outputs, and output-level proxies risk circularity when used to calibrate the very metrics designed to go beyond them.

\paragraph{Bogus data as a proxy for not knowing.}
We sidestep this measurement problem by grounding knowledge in dataset construction rather than model training. As the negative class, we draw from data. For example, the TOFU benchmark~\citep{maini_tofu_2024} consists entirely of fictitious authors invented for the benchmark and absent from any pre-training corpus. Because these entities were never written about anywhere, no model can have parametric knowledge of them, and there is no plausible mechanism by which such knowledge could exist. As the positive class, we can draw from Harry Potter (HP) QA pairs; HP knowledge is confirmed to persist in the parameters of most LLMs as a consequence of large-scale pre-training~\citep{carlini_extracting_2021,seyitoglu_extracting_2024}.
A metric that separates HP from TOFU under this protocol has therefore demonstrated genuine knowledge detection, not behavioral correlation. The specific datasets and model counts used in our evaluation are described in Section~\ref{sec:experiments}.

%% file: Sections/04-expirements.tex
\section{Experiments}
\label{sec:experiments}

\begin{enumerate}
    \item RQ1: How can we measure the binary existence of knowledge in LLMs? Can we suggest a metric that maximizes separation between known and unknown data?
    \item RQ2: How can we measure fine-grained knowledge extraction ability from LLMs? Which metric has the highest correlation with Attack success rate (ASR)?
\end{enumerate}

\subsection{RQ1: Fragmentation as a Proxy for Parametric Knowledge}

\paragraph{Setup.}
We evaluate all nine fragmentation metrics under the protocol presented in (Section~\ref{sec:semantic_testing}) across ten instruction-tuned LLMs from three families: Qwen ($\times$4), Gemma ($\times$2), and LLaMA ($\times$4), ranging from 7B to 14B parameters (Appendix~\ref{sec:impl_details}), using batch size $B{=}10$ (full sensitivity sweep over $B\in\{1,2,5,10,20,50\}$ justifying this choice in Appendix~\ref{sec:impl_anon}).
The known-concept class consists of 96 QA pairs drawn from the Harry Potter novels; the unknown-concept class consists of 100 QA pairs from the TOFU benchmark~\citep{maini_tofu_2024}.
All ten models score 0\% QA accuracy on TOFU, casually confirming that no knowledge of these fictitious entities exists.
For each model, fragmentation is computed at the per-metric layer that maximizes HP--TOFU separation. Appendix~\ref{sec:layer_selection} also reports a fixed, deployable heuristic, the layer at $0.8\times$ relative model depth, requiring no per-model search, that matches the oracle sweep's discriminability.

\paragraph{Results.}
\textbf{All nine metrics correctly separate HP from TOFU across all 10 models and survive BH  (Benjamini-Hochberg false discovery rate control at $\alpha=0.05$) correction} 
(Table~\ref{tab:rq1_disc}; per-model raw values in Appendix Tables~\ref{tab:rq1_main}-\ref{tab:rq1_entropy}).
AUSS$_{\cos^2}$ achieves the largest separation ($|d|{=}21.51$, $p_\text{adj}{=}0.002$) and AUSS$_\text{L2}$ is second ($|d|{=}18.91$), both dispersion higher for TOFU than HP in 10/10 models; the scale-invariant classifier check below (AUC$=1.0$) confirms this separation holds independent of each metric's raw scale.

Entropy-based dispersion metrics (NormAnonDiff~$S_2$, AnonDiff~$S_2$, Reg~$S_2$; Section~\ref{sec:metrics}) likewise reach 10/10 with large effect sizes.
Coherence metrics, i.e. centroid norm $\|\bar{\Delta\hat{v}}\|$, Reg~$\overline{\cos}$, AnonDiff~$\overline{\cos}$, and AnonDiff~$\bar{L}_2$, all show strictly higher values for HP (8/10 or 10/10 models), confirming that a known concept induces a tighter, more directionally consistent hidden-state geometry than a purely fictional one.
Effect sizes reflect per-metric peak-layer selection; performance under a label-free layer protocol would be lower (Appendix~\ref{sec:layer_selection}).

\textbf{The separation generalizes beyond a single narrow contrast and is not a scale artifact.}
To rule out that this separation reflects one narrow contrast, we extend the comparison to Harry Potter (retain split) and Star Wars against the same TOFU baseline; all 24 BH-corrected comparisons across the 8 AUSS metrics are significant ($q\in[0.0063,0.0260]$), with HP-retain effect sizes reaching $d{=}6.20$, larger than the original HP-forget contrast (Appendix Table~\ref{tab:fourway}).
A scale-invariant leave-one-model-out classifier trained on the standardized 8-metric peak-layer feature vector, evaluated on the original HP-forget vs.\ TOFU pair, confirms the known/unknown separation is not an artifact of raw effect-size scale: Logistic Regression reaches 100\% accuracy (AUC$=1.0$, 9/9 models fully correct) on the fresh, direction-corrected extraction (Appendix Table~\ref{tab:classifier}).
\begin{table}[t]
\centering
\small
\setlength{\tabcolsep}{4pt}
\caption{
    \textbf{RQ1: Metric discriminability} (HP known vs.\ TOFU unknown) across 10 LLMs, evaluated at the per-metric layer maximising HP--TOFU separation, sorted by Cohen's $|d|$.
    Gap: mean $|\text{HP}{-}\text{TOFU}|$ with 95\% bootstrap CI.
    $p_\text{adj}$: Wilcoxon signed-rank, BH-corrected ($n{=}10$); ${}^{**}p{<}0.01$.
    ${}^\star$: to concept-differential vectors.
    ${}^\dagger$: raw-activation baseline, no anon-vector subtraction.
    Per-model values in Appendix (Tables~\ref{tab:rq1_main}--\ref{tab:rq1_entropy}).
}
\label{tab:rq1_disc}
\begin{tabular}{|l|ccc|}
\hline
Metric & $|\delta|$ & Gap [95\% CI] & $p_\text{adj}$ \\
\hline\hline
\multicolumn{4}{|l|}{\textit{Dispersion ($\downarrow$ HP expected; 10/10 models)}} \\
\hline
AUSS$_{\cos^2}$                  & $21.51$ & $0.686$ $[0.667,\,0.705]$ & $0.002^{**}$ \\
AUSS$_\text{L2}$                 & $18.91$ & $1.203$ $[1.166,\,1.239]$ & $0.002^{**}$ \\
NormAnonDiff $S_2^\star$         & $4.21$  & $1.206$ $[1.022,\,1.332]$ & $0.002^{**}$ \\
AnonDiff $S_2^\star$             & $3.50$  & $0.804$ $[0.671,\,0.941]$ & $0.002^{**}$ \\
Reg $S_2^\dagger$                & $0.92$  & $0.117$ $[0.054,\,0.200]$ & $0.002^{**}$ \\
\hline
\multicolumn{4}{|l|}{\textit{Coherence ($\uparrow$ HP expected)}} \\
\hline
$\|\bar{\Delta\hat{v}}\|$        & $1.69$ & $0.182$ $[0.118,\,0.244]$ & $0.002^{**}$ \\
AnonDiff $\overline{\cos}$       & $1.29$ & $0.128$ $[0.069,\,0.185]$ & $0.010^{**}$ \\
Reg $\overline{\cos}^\dagger$    & $1.25$ & $0.044$ $[0.023,\,0.065]$ & $0.002^{**}$ \\
AnonDiff $\bar{L}_2$             & $1.21$ & $2.634$ $[1.412,\,3.946]$ & $0.002^{**}$ \\
\hline
\end{tabular}
\end{table}

\subsection{RQ2: Fragmentation as a Predictor of Unlearning Robustness}

\paragraph{Setup.}
RQ1 established that fragmentation geometry reflects whether knowledge exists. RQ2 asks whether it also reflects how \emph{recoverable} that knowledge remains after unlearning. Fragmentation metrics for each model--concept--method triple are computed at the layer that minimises $S_2$ over the post-unlearning forget-set queries (label-free protocol; Appendix~\ref{sec:layer_selection}).
The layer-resolved sweep evaluates every transformer layer independently (Appendix~\ref{sec:layer_analysis}).

\paragraph{Models and unlearning methods.}
We apply five unlearning methods spanning output-level suppression to representation-level distortion:
\textbf{GA}~\cite{yao_large_2024} negates the forget-set CE loss;
\textbf{DPO}~\cite{rafailov_direct_2024} treats forget pairs as rejected and retain pairs as chosen;
\textbf{NPO}~\cite{zhang_negative_2024} applies a negative preference loss against a frozen reference;
\textbf{NPO+KL}~\cite{zhang_negative_2024} augments NPO with KL retain regularisation;
and \textbf{RMU}~\cite{huu-tien_effects_2025} steers forget-set hidden states at a fixed intermediate layer toward a random target direction.
GA and DPO act primarily on outputs, whereas RMU explicitly distorts internal representations. This distinction is reliably captured by fragmentation metrics even when the two approaches yield indistinguishable ROUGE-L scores (Appendix~\ref{sec:impl_unlearn}). 
These methods are applied to three instruction-tuned models: LLaMA-2-7B-chat, Meta-LLaMA-3-8B-Instruct, and Qwen2.5-7B-Instruct, each also evaluated as an unmodified \textbf{Base} control, across three target concepts: \textit{Harry Potter} \cite{eldan_whos_nodate}, \textit{Star Wars}, and \textit{William Shakespeare} \cite{hong_intrinsic_nodate} ($3{\times}6{\times}3{=}54$ conditions; incomplete runs shown as ``---'').
A run is verified ($\checkmark$) when Forget ROUGE-L $<0.10$ and Retain ROUGE-L $\geq 0.80\times$ baseline; surface suppression alone does not guarantee robustness.
Four attacks then probe recoverability: Activation Steering~\cite{seyitoglu_extracting_2024}, ICL~\cite{brown_language_2020}, GCG~\cite{zou_universal_2023}, and MIA~\cite{fu_membership_nodate} at geometric, contextual, gradient, and distributional levels, respectively; lower values indicate stronger unlearning (Appendix~\ref{sec:impl_attacks}).
Correlation analysis excluding Base and WHP (Who's Harry Potter; \citealt{eldan_whos_nodate}) runs ($n{=}44$--$45$ non-base runs with complete attack data).

\paragraph{Results.}
Figure~\ref{fig:rq2_bar} summarizes attack success by method across all three models; full per-run values are in Appendix~\ref{sec:rq2_full_runs} (Tables~\ref{tab:rq2_main_llama2}--\ref{tab:rq2_main_qwen}).
Table~\ref{tab:rq2_best} shows every BH-significant geometry predictor per attack ($n{=}44$--$45$ non-Base, non-WHP runs); full Spearman table in Appendix Table~\ref{tab:rq2_corr}.

\textbf{AUSS fragmentation geometry predicts ICL recoverability; entropy-family metrics predict GCG and Steering.}
ICL is the most significantly predictable attack via the AUSS family: AUSS$_\text{L2}$ ($\rho{=}{-}0.51^{**}$), centroid norm ($\rho{=}{+}0.51^{**}$), and AUSS$_{\cos^2}$ ($\rho{=}{-}0.50^{**}$) all survive BH correction; methods that geometrically fragment hidden-state representations more thoroughly are harder to reach via in-context priming.
GCG and Steering are predictable in aggregate via the Gram-entropy (Reg) and AnonDiff families rather than AUSS: Reg~$\overline{\cos}$ ($\rho{=}{-}0.51^{**}$) and Reg~$S_2$ ($\rho{=}{+}0.45^{**}$) predict GCG, and AnonDiff~$S_2$ ($\rho{=}{+}0.48^{**}$) predicts Steering.
MIA is omitted: no metric survives BH correction (Appendix Table~\ref{tab:rq2_corr}).
Leave-one-run-out validation across verified unlearning runs ($n{=}26$--$27$) confirms ICL is the only attack that is both leave-one-out stable (8/8 metrics) and BH-significant (7/8); Steering (8/8) and GCG (7/8) are largely stable but not significant, while MIA (2/8) is neither reliably stable nor significant (Appendix~\ref{sec:loo_stability}).

\begin{figure*}[!ht]
\centering
\includegraphics[width=\textwidth]{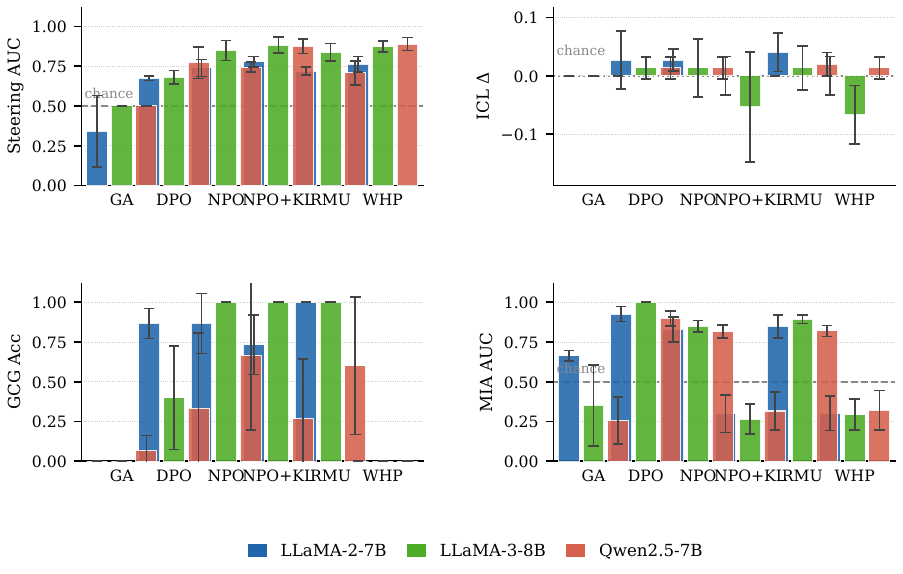}
\caption{
    \textbf{RQ2: Mean attack success per unlearning method} across four attack types (2$\times$2 grid), averaged over three concepts (error bars~= std).
    Bars are grouped by unlearning method; colour indicates model (LLaMA-2-7B: blue, LLaMA-3-8B: green, Qwen2.5-7B: orange-red).
    Dashed lines mark chance level (0.5 for AUC-based attacks; 0 for ICL~$\Delta$).
    \textbf{GA} is the only method that consistently drives Steer-AUC near chance across all three models.
    \textbf{LLaMA-3} (green) reaches GCG Acc~$=0$ under every method, revealing strong model-family variation in gradient-attack vulnerability.
    ICL~$\Delta$ is near zero across all methods and models; the geometric correlation with ICL recoverability emerges from the layer-resolved analysis (Table~\ref{tab:layer_best}), not from this aggregate view.
    Full per-run values in Appendix~\ref{sec:full_results} (Tables~\ref{tab:rq2_main_llama2}--\ref{tab:rq2_main_qwen}).
}
\label{fig:rq2_bar}
\end{figure*}

\begin{table}[t]
\centering
\small
\setlength{\tabcolsep}{5pt}
\caption{
    \textbf{RQ2: BH-significant geometry predictors per attack} (Spearman $\rho$, $n{=}44$--$45$ non-Base non-WHP runs with complete attack data, BH-corrected within each attack's 9-metric column).
    All metrics shown survive BH at $q{<}0.05$; full 9-metric table (all attacks, including non-significant cells) in Appendix Table~\ref{tab:rq2_corr}.
}
\label{tab:rq2_best}
\begin{tabular}{|l|c|c|c|}
\hline
Metric & ICL & GCG & Steering \\
\hline\hline
AUSS$_\text{L2}$              & $-0.51^{**}$        & --            & -- \\
AUSS$_{\cos^2}$                & $-0.50^{**}$        & --            & -- \\
$\|\bar{\Delta\hat{v}}\|$      & $+0.51^{**\dagger}$ & --            & -- \\
Reg~$\overline{\cos}$          & --                  & $-0.51^{**}$  & -- \\
Reg~$S_2$                      & --                  & $+0.45^{**}$  & -- \\
AnonDiff~$S_2$                  & --                  & --            & $+0.48^{**}$ \\
\hline
\end{tabular}
\end{table}
${}^\dagger$$\|\bar{\Delta\hat{v}}\|$ is the exact algebraic dual of AUSS$_\text{L2}$ (identical $|\rho|$, opposite sign by construction, since AUSS$_\text{L2}{=}2{-}2\bar c$ for mean pairwise cosine $\bar c$ and $\|\bar{\Delta\hat{v}}\|$ is a monotonic function of the same $\bar c$), not an independent confirmation.

\begin{figure}[t]
\centering
\includegraphics[width=\columnwidth]{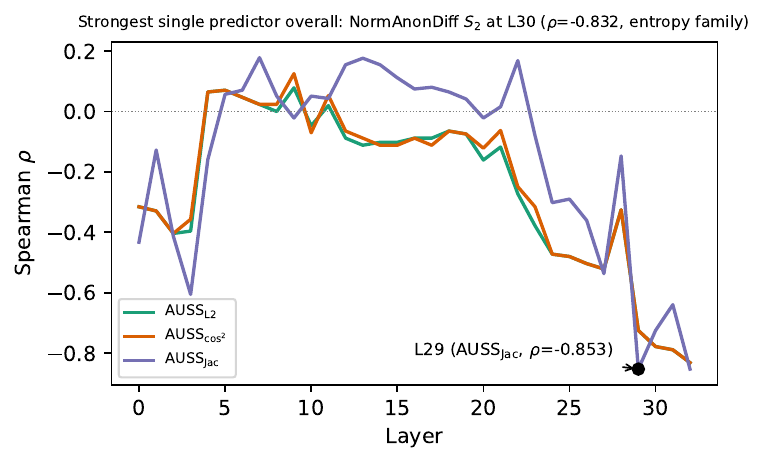}
\caption{
    \textbf{Spearman $\rho$ by layer, GCG attack.}
    Each line is one AUSS fragmentation metric; the peak at L29 (AUSS$_\text{Jac}$, $\rho{=}{-}0.853$) is marked.
    The strongest single predictor overall is NormAnonDiff~$S_2$ at L30 ($\rho{=}{-}0.832$, entropy family; Appendix~\ref{sec:layer_analysis}).
    GCG recoverability is governed by late-layer geometry, close to where ICL's own peak falls (Appendix~\ref{sec:layer_analysis}).
    (Full four-attack sweep in Appendix~\ref{sec:layer_analysis}.)
}
\label{fig:layer_rho_gcg}
\end{figure}

\paragraph{Layer-resolved geometry concentrates late in the network for the two attacks it reliably predicts.}
The aggregate correlations in Table~\ref{tab:rq2_best} reflect peak-layer values averaged over all metric-layer combinations.
When we instead sweep Spearman~$\rho$ across \emph{every} transformer layer for both AUSS and entropy metrics (Figure~\ref{fig:layer_rho_gcg}; full results in Appendix~\ref{sec:layer_analysis}), only GCG and ICL survive global BH correction, and both peak in the same late-network region:

\textbf{GCG}: AUSS$_\text{Jac}$ at \textbf{L29} reaches $\rho{=}{-}0.853$ ($p{=}0.0008$, $n{=}11$ runs with complete layer-resolved data, smaller than the $n{=}44$--$45$ aggregate pool), with NormAnonDiff~$S_2$ close behind at L30 ($\rho{=}{-}0.832$, $p{=}0.0015$); the gradient-based suffix attack's signal is shaped by whichever layer the model's final next-token decision is most sensitive to.

\textbf{ICL}: Batch\_Cov\_TopEig at \textbf{L28} reaches $\rho{=}{-}0.886$ ($p{<}0.0001$, $n{=}15$), one layer from GCG's own peak; in-context priming must similarly override the model's late-stage, generation-ready representation.

\textbf{MIA} and \textbf{Steering} show only weak, non-BH-significant layer-specific signal in the current dataset (best entropy-family points at L31--L32, $\rho{\approx}0.55$--$0.65$, uncorrected $p{<}0.05$ but $q_\text{global}{\geq}0.22$); we report these descriptively rather than as reliable predictors (Appendix Table~\ref{tab:layer_best}).

The highest-performing layer is highly predictable across models, consistently residing within the $0.80$--$0.88$ range of total model depth. A single fixed relative depth of $0.8\times$ model depth, with no per-model information at all, already gives 8/8 metrics correct-direction and significant at $n{=}9$; leave-one-model-out transfer of this depth achieves 72/72 (model, metric) pairs correct-direction, identical to an oracle per-model search (Appendix~\ref{sec:layer_selection}).

\subsection{Benchmarking Against Prior Intrinsic Probes}

\paragraph{Trivial raw-similarity baselines.}
Table~\ref{tab:rq2_corr} already includes head-to-head comparisons against trivial intrinsic baselines: the Reg~$S_2$ (adapted from \citealt{skean_layer_2025}) and Reg~$\overline{\cos}$ rows measure raw, non-anonymised hidden states with no anon-vector subtraction.

\paragraph{Head-to-head against published method.}
We benchmark AUSS directly against the representational metrics of \citet{xu_unlearning_2025} (PCA-similarity, PCA-shift, CKA), which, unlike our method, require both pre- and post-unlearning model checkpoints. Across all 54 RQ2 configurations, BH correction over the full 44-test pool shows a mixed, attack-dependent picture: AUSS's best metric outperforms Xu et al.'s best metric on ICL ($\rho{=}{-}0.572$, $q{=}0.0001$ vs.\ $\rho{=}{+}0.381$, $q{=}0.0212$) and Steering ($\rho{=}{-}0.452$, $q{=}0.0042$ vs.\ $\rho{=}{+}0.428$, $q{=}0.0076$), while Xu et al.'s CKA outperforms AUSS's best metric on GCG ($\rho{=}{+}0.374$, $q{=}0.0212$ vs.\ $\rho{=}{-}0.315$, $q{=}0.068$, not BH-significant) and MIA (neither survives correction: $\rho{=}{+}0.336$ vs.\ $\rho{=}{+}0.085$).
11 of 44 tests survive BH correction; 8 of those 11 are AUSS metrics (full 44-test table in Appendix Table~\ref{tab:xu_comparison}).

%% file: Sections/05-discussion.tex
\section{Discussion}
\label{sec:discussion}

\paragraph{The geometry of parametric knowledge.}
RQ1 establishes that activation fragmentation reliably distinguishes known from unknown
concepts, and the geometry explains why.
The anon-vector construction subtracts the mean hidden state of identity-stripped
paraphrases, canceling the shared linguistic and syntactic signal and leaving a residual
that reflects only the concept-specific content encoded at that layer.
When a model parametrically knows a concept, its internal representations across distinct
queries about that concept agree on a direction the residual vectors form a tight,
low-entropy cluster.
When the concept is absent from the parameters, no such agreement exists: the residual
after syntax cancellation is noise, and the vectors scatter into high-variance fragmented
directions.
This geometric signature is what our metrics quantify, and it is entirely invisible to
behavioral evaluation.

\paragraph{Architecture-agnostic universality.}
The coherence signature described above is not an artifact of any particular architecture.
The 10/10 correct-direction result for the strongest metrics
(AUSS$_{\cos^2}$, AUSS$_{L_2}$, AnonDiff~$S_2$) in RQ1 spans three model families
(Qwen, Gemma, and LLaMA) across different architectures, training recipes, and parameter
counts from 2B to 14B.
This cross-architecture consistency suggests that the geometric signature of parametric
knowledge is not idiosyncratic to any particular model design.
Fragmentation metrics qualify as model-family-neutral auditing tools, requiring
no architecture-specific calibration.

\paragraph{Suppression versus deletion.}
A consistent RQ2 pattern is that reducing Forget ROUGE-L does not reliably reduce
fragmentation: GA applied to LLaMA-3-8B reaches Forget ROUGE-L $=0.000$ yet leaves
AUSS$_{L_2}$ and $\|\bar{\Delta v}\|$ nearly unchanged from the Base model.
GA and DPO act primarily on the output distribution, leaving internal concept geometry
largely intact, whereas RMU, which steers representations directly, shifts fragmentation
more noticeably but does not erase it.
This matches the redirection-rather-than-deletion picture reported by
\citet{xu_unlearning_2025} and \citet{schoepf_redirection_2026}: representation-level
interventions come closest to genuine deletion, and fragmentation metrics measure how far
even they fall short.

\paragraph{Layer-resolved geometry concentrates late in the residual stream for the attacks it reliably predicts.}
Of the four attacks, GCG and ICL are the two whose layer-specific signal survives BH
correction at the global scope (Appendix~\ref{sec:layer_analysis}), and both concentrate
their strongest signal in the same late-network region: GCG peaks at L29
(AUSS$_\text{Jac}$, $\rho{=}{-}0.85$) and ICL at L28 (Batch\_Cov\_TopEig, $\rho{=}{-}0.89$),
one layer apart.
This is mechanistically coherent rather than coincidental.
GCG's adversarial suffix is optimised end-to-end by backpropagating through the entire
network, so its gradient signal is shaped by whichever layer the model's final next-token
decision is most sensitive to.
ICL must similarly override the model's late-stage, generation-ready representation to
redirect its output.
Both attacks are therefore best diagnosed by measuring fragmentation deep in the residual
stream rather than at an early or intermediate depth.
MIA and Steering show only weak, non-BH-significant layer-specific signal in the current
dataset (best entropy-family points at L31--L32, $\rho{\approx}0.55$--$0.65$, uncorrected
$p{<}0.05$ but not surviving global correction); we treat these descriptively rather than
as reliable single-layer predictors.
Appendix~\ref{sec:layer_selection} reports a practical, label-free layer-selection rule
that performs close to this attack-specific oracle search without per-attack calibration.

\paragraph{Attack-dependent sign phenomena.}
The correlation's sign is not uniform across attacks (Table~\ref{tab:rq2_corr}):
AUSS$_{L_2}$ predicts ICL negatively ($\rho{=}{-}0.51$), while Reg~$S_2$ and
AnonDiff~$S_2$ predict GCG and Steering positively ($\rho{=}{+}0.45$, $\rho{=}{+}0.48$).
Based on the different mechanisms by which these attacks recover knowledge, we expect
their correlation with fragmentation to differ in signal.
ICL recovers "knowledge" through the model's own generation, which plausibly requires a
single coherent direction, empirically it seems that fragmentation tracks the absence of one.
GCG and Steering instead work by searching or intervening on the
representation directly, where possibly a more dispersed representation offers more directions
to exploit, which would explain the opposite sign.
We discuss in Limitations how this account could be further validated in future work.

\paragraph{Practical implications for unlearning evaluation.}
Activation fragmentation can serve as a computationally cheap, pre-attack screening step
in unlearning evaluation pipelines.
If fragmentation metrics confirm that the concept geometry has been genuinely disrupted,
practitioners can have higher confidence that the unlearning is not merely suppression;
if fragmentation remains intact, that is a strong signal that adversarial recovery will
succeed without running expensive attacks.
Its forward-pass-only design makes it applicable even where generation is prohibited,
such as API-access audits subject to output restrictions.

%% file: Sections/06-related.tex
\section{Related Work}
\label{sec:related}

\paragraph{Parametric Knowledge and Intrinsic Evaluation.}
Factual associations are geometrically structured within intermediate transformer layers \cite{petroni_language_2019,geva_transformer_2021,meng_locating_2023}. While recent intrinsic probes aim to evaluate this knowledge without token generation \cite{gottesman_estimating_2024,hong_intrinsic_nodate}, they frequently conflate conceptual content with query syntax or require a reference model. We build upon the Gram-matrix entropy framework \cite{skean_layer_2025} and the geometric grounding of intermediate layers \cite{razzhigaev_shape_2023}, but depart from prior work by measuring the \emph{population geometry} of anon-vectors. This isolates conceptual identity from structural artifacts without relying on pre-unlearning baselines.

\paragraph{Machine Unlearning and Adversarial Robustness.}
Current unlearning interventions---spanning gradient ascent, preference optimisation, and representation steering \cite{yao_large_2024,rafailov_direct_2024,huu-tien_effects_2025}---frequently produce \emph{suppression} rather than genuine deletion \cite{xu_unlearning_2025,lee_mechanistic_2024,schoepf_redirection_2026}. Because knowledge remains structurally intact, adversarial techniques such as activation steering \cite{seyitoglu_extracting_2024}, in-context learning \cite{brown_language_2020}, and GCG \cite{zou_universal_2023} can bypass behavioral safeguards to recover suppressed facts. Prior representation-level audits either require pre- and post-unlearning model comparisons \cite{xu_unlearning_2025} or the active execution of these costly attacks \cite{lynch_eight_2024}. In contrast, our fragmentation framework offers a \emph{single-phase}, generation-free proxy designed to \emph{predict} vulnerability to these exact recovery channels using a single geometric forward pass.

%% file: Sections/07-Limitations.tex
\section{Limitations}
\label{sec:limitations}
Our evaluation has several limitations. First, deterministic string substitution for anonymisation risks tokenizer-driven lexical artifacts. We mitigate this using morphologically similar names and multi-variant averaging ($K=5$) in RQ2 (supported by consistent ablations in Appendix~\ref{sec:impl_anon}), though future work should explore tokenization-matched replacements, LLM paraphrasing, or systematically benchmarking alternative anonymisation pipelines beyond the scheme used here. Second, relying on a single anonymised variant ($K=1$) per query in RQ1 may retain idiosyncratic phrasing, which we offset by averaging across $N=96$--$100$ queries per concept. Third, our batch-based construction requires $B\geq2$ forget-set samples to define a valid direction vector, which may not suit single-sample or very small-sample unlearning settings; a promising extension augments a single target sample with neutral or synthetic data to compute a pseudo-batch direction, which we leave to future work. 
Fourth, the explanation offered in Section~\ref{sec:discussion} for why fragmentation predicts ICL and GCG/Steering with opposite sign is inferred from the correlational pattern alone; we do not validate it against the models' actual generations or intermediate reasoning traces, and confirming it would require inspecting chain-of-thought or generation-level outputs alongside the geometry, which we leave to future work. Finally, our concept coverage in RQ1, while now validated across four conditions (Harry Potter forget/retain, Star Wars, TOFU; Appendix~\ref{sec:rq1_multidomain}), remains limited to English-language fictional/factual domains. Expanding to graded knowledge levels and non-fiction technical domains is necessary to fully confirm the generality of our findings.

%% file: Sections/08-appendix.tex
\appendix

\section{Mathematical Proofs}
\label{sec:appendix}
\label{app:proofs}

This section provides complete proofs for the theoretical results stated in
Section~\ref{sec:mathematics}.

\subsection{Proof of Theorem~\ref{thm:entropy_fragmentation}}

\begin{theorem*}[High entropy forces geometric fragmentation]
Let $A \in \mathbb{R}^{n \times d}$ be a matrix of $L_2$-normalised activation
vectors and let $G = AA^\top$ be the corresponding Gram matrix with
$\mathrm{EffRank}(G) = \mathrm{Tr}(G)/\lambda_{\max}(G)$.
Then the mean pairwise squared Euclidean distance satisfies
\[
  D_{L_2}^2 \;\geq\; 2 - \frac{2}{\mathrm{EffRank}(G)}.
\]
\end{theorem*}

\begin{proof}
Let $v_1,\ldots,v_n \in \mathbb{R}^d$ denote the rows of $A$, with
$\|v_i\|_2 = 1$ for all $i$.

\medskip
\noindent\textbf{Step 1 (Gram matrix structure).}
Because the rows are unit vectors, $G_{ij} = v_i^\top v_j$ and the diagonal
satisfies $G_{ii} = 1$, so $\mathrm{Tr}(G) = n$.
Hence $\mathrm{EffRank}(G) = n/\lambda_{\max}(G)$.

\medskip
\noindent\textbf{Step 2 (Distance--cosine identity).}
For any pair $i \neq j$,
\[
  \|v_i - v_j\|_2^2
  = \|v_i\|_2^2 - 2v_i^\top v_j + \|v_j\|_2^2
  = 2 - 2G_{ij}.
\]

\medskip
\noindent\textbf{Step 3 (Express $D_{L_2}^2$ via the Gram matrix).}
Define the mean off-diagonal cosine similarity
\[
  \bar{c} \;=\; \frac{1}{n(n-1)}\sum_{i \neq j} G_{ij},
\]
so that $D_{L_2}^2 = 2 - 2\bar{c}$.
Summing all entries of $G$ and separating diagonal from off-diagonal,
\[
  \mathbf{1}^\top G\,\mathbf{1}
  = \mathrm{Tr}(G) + \sum_{i \neq j}G_{ij}
  = n + n(n{-}1)\bar{c}.
\]
Solving for $\bar{c}$ and substituting:
\begin{align*}
  D_{L_2}^2
  &= 2 - 2\bar{c} \\
  &= \frac{2n - 2\,\mathbf{1}^\top G\,\mathbf{1}/n}{n-1}.
\end{align*}

\medskip
\noindent\textbf{Step 4 (Rayleigh quotient bound).}
For any unit vector $x$, $\lambda_{\max}(G) \geq x^\top G x$.
Taking $x = \mathbf{1}/\sqrt{n}$ (which satisfies $\|x\|_2 = 1$):
\[
  \lambda_{\max}(G) \;\geq\; \frac{\mathbf{1}^\top G\,\mathbf{1}}{n}.
\]

\medskip
\noindent\textbf{Step 5 (Conclusion).}
From Step 4, $\mathbf{1}^\top G\,\mathbf{1}/n \leq \lambda_{\max}(G)$,
so $2n - 2\,\mathbf{1}^\top G\,\mathbf{1}/n \geq 2n - 2\lambda_{\max}(G)$.
Dividing by $n-1 > 0$:
\begin{align*}
  D_{L_2}^2
  &= \frac{2n - 2\,\mathbf{1}^\top G\,\mathbf{1}/n}{n-1}
  \;\geq\; \frac{2(n - \lambda_{\max}(G))}{n-1} \\
  &= \frac{2n}{n-1}\!\left(1 - \frac{1}{\mathrm{EffRank}(G)}\right).
\end{align*}
Since $n/(n-1) \geq 1$ for all $n \geq 2$,
\begin{align*}
  D_{L_2}^2
  &\geq 2\!\left(1 - \frac{1}{\mathrm{EffRank}(G)}\right) \\
  &= 2 - \frac{2}{\mathrm{EffRank}(G)}.
\end{align*}
\end{proof}

\begin{remark}
The proof actually establishes the tighter bound
$D_{L_2}^2 \geq \tfrac{2n}{n{-}1}\bigl(1 - 1/\mathrm{EffRank}(G)\bigr)$;
the theorem states the slightly weaker form $2 - 2/\mathrm{EffRank}(G)$,
which is independent of $n$ and easier to interpret.
The bound is tight (up to the $n/(n{-}1)$ factor) when all $n$ vectors
are mutually orthogonal: then $G = I_n$, $\lambda_{\max}(G) = 1$,
$\mathrm{EffRank}(G) = n$, and $D_{L_2}^2 = 2 = 2 - 2/n + 2/n$.
\end{remark}

\subsection{Entropy--EffRank Relationship}
\label{app:entropy_effrank}

The following proposition formalises the claim in
Section~\ref{sec:mathematics} that a high Rényi-2 Gram-matrix entropy
$S_2$ forces a large effective rank, completing the implication chain
$\text{high }S_2 \Rightarrow \text{high EffRank} \Rightarrow \text{large }D_{L_2}^2$.

\begin{proposition}[Entropy--EffRank sandwich]
\label{prop:entropy_effrank}
Let $G$ be a positive semi-definite matrix with normalised spectral
distribution $p_i = \lambda_i(G)/\mathrm{Tr}(G) \geq 0$, $\sum_i p_i = 1$,
ordered $p_1 \geq p_2 \geq \cdots \geq 0$.
Define $S_2(G) = -\log_2\!\bigl(\sum_i p_i^2\bigr)$ and
$\mathrm{EffRank}(G) = 1/p_1$.  Then
\[
  2^{S_2(G)/2} \;\leq\; \mathrm{EffRank}(G) \;\leq\; 2^{S_2(G)}.
\]
\end{proposition}

\begin{proof}
\textbf{Upper bound.}
Because $p_1 \geq p_i$ for every $i$,
\[
  \sum_i p_i^2 \;\leq\; p_1\sum_i p_i = p_1.
\]
Taking reciprocals (both sides positive) reverses the inequality:
$1/p_1 \leq 1/\sum_i p_i^2$,
i.e., $\mathrm{EffRank}(G) \leq 2^{S_2(G)}$.

\medskip
\noindent\textbf{Lower bound.}
Trivially, $p_1^2 \leq \sum_i p_i^2$ (one non-negative term cannot exceed their sum).
Taking square roots: $p_1 \leq \sqrt{\sum_i p_i^2} = 2^{-S_2(G)/2}$.
Hence $\mathrm{EffRank}(G) = 1/p_1 \geq 2^{S_2(G)/2}$.
\end{proof}

\begin{corollary}[High $S_2$ forces geometric fragmentation]
\label{cor:s2_fragmentation}
For any matrix $A \in \mathbb{R}^{n \times d}$ of $L_2$-normalised vectors
with Gram matrix $G = AA^\top$,
\[
  D_{L_2}^2
  \;\geq\; 2 - 2\cdot 2^{-S_2(G)/2}
  \;=\; 2\bigl(1 - 2^{-S_2(G)/2}\bigr).
\]
In particular, $D_{L_2}^2 \to 2$ as $S_2(G) \to \infty$.
\end{corollary}

\begin{proof}
From Proposition~\ref{prop:entropy_effrank},
$\mathrm{EffRank}(G) \geq 2^{S_2(G)/2}$, so
$1/\mathrm{EffRank}(G) \leq 2^{-S_2(G)/2}$.
Substituting into Theorem~\ref{thm:entropy_fragmentation}:
$D_{L_2}^2 \geq 2 - 2/\mathrm{EffRank}(G) \geq 2 - 2\cdot 2^{-S_2(G)/2}$.
\end{proof}

\begin{remark}
Proposition~\ref{prop:entropy_effrank} shows that $S_2$ and
$\mathrm{EffRank}$ are tightly coupled: both grow together as the
spectral mass spreads across more eigenvalue directions.
The upper bound $\mathrm{EffRank} \leq 2^{S_2}$ also shows that the
two quantities agree to within a factor of two on the log-scale,
justifying the shorthand ``$S_2$ is an increasing function of
$\mathrm{EffRank}$'' used in the main text.
\end{remark}

\subsection{Proof of Theorem~\ref{thm:clustering} and the Exact Entropy Identity}
\label{app:clustering_proof}

We first establish a key lemma that is an exact algebraic identity—no
inequalities, no extra assumptions.
It shows that $S_2$ and the mean squared pairwise cosine are two
encodings of the same spectral quantity.

\begin{lemma}[Gram entropy equals mean squared cosine]
\label{lem:s2_cos2_identity}
Let $v_1,\ldots,v_n \in \mathbb{R}^d$ be $L_2$-normalised vectors with
Gram matrix $G = AA^\top$ and normalised Gram $K_G = G/\mathrm{Tr}(G) = G/n$.
Let $\overline{\cos^2} = \frac{1}{n(n-1)}\sum_{i\neq j}(v_i^\top v_j)^2$
be the mean pairwise squared cosine similarity.  Then
\begin{equation}
  \mathrm{Tr}(K_G^2) \;=\; \frac{1 + (n-1)\,\overline{\cos^2}}{n}
  \label{eq:s2_cos2_identity}
\end{equation}
equivalently $S_2(G) = -\log_2\!\left(\tfrac{1+(n-1)\overline{\cos^2}}{n}\right)$.
In particular, $S_2$ is a \emph{strictly decreasing bijection} of
$\overline{\cos^2}$: $S_2 = 0 \Leftrightarrow \overline{\cos^2} = 1$
(all pairs parallel or anti-parallel), and $S_2 = \log_2 n \Leftrightarrow
\overline{\cos^2} = 0$ (all pairs orthogonal).
\end{lemma}

\begin{proof}
Expanding the Frobenius norm of $G$:
\begin{align*}
  \mathrm{Tr}(G^2) &= \sum_{i,j=1}^n G_{ij}^2 \\
  &= \underbrace{\sum_i G_{ii}^2}_{=\,n \text{ (diagonal }= 1)} + \sum_{i\neq j}G_{ij}^2 \\
  &= n + n(n{-}1)\,\overline{\cos^2}.
\end{align*}
Dividing by $\mathrm{Tr}(G)^2 = n^2$:
\begin{align*}
  \mathrm{Tr}(K_G^2)
  &= \frac{\mathrm{Tr}(G^2)}{n^2}
  = \frac{n + n(n{-}1)\,\overline{\cos^2}}{n^2} \\
  &= \frac{1 + (n{-}1)\,\overline{\cos^2}}{n}.
\end{align*}
Since $\overline{\cos^2} \in [0,1]$ (each $G_{ij} \in [-1,1]$), the
right-hand side is continuous and strictly increasing in $\overline{\cos^2}$,
making $S_2 = -\log_2(\cdot)$ a strictly decreasing bijection.
Boundary values: $\overline{\cos^2} = 1$ gives $\mathrm{Tr}(K_G^2) = 1$,
$S_2 = 0$; $\overline{\cos^2} = 0$ gives $\mathrm{Tr}(K_G^2) = 1/n$,
$S_2 = \log_2 n$.
\end{proof}

\begin{remark}
Lemma~\ref{lem:s2_cos2_identity} gives the converse of
Theorem~\ref{thm:entropy_fragmentation} \emph{for the squared-cosine
metric}: $S_2 \to 0$ if and only if $\overline{\cos^2} \to 1$, which
means every pair of vectors satisfies $|v_i^\top v_j| \to 1$, i.e.\
they are either perfectly aligned or perfectly anti-aligned.
Crucially, a low $S_2$ alone does not guarantee that vectors point in the
\emph{same} direction (positive cosines); two anti-parallel unit vectors
achieve $S_2 = 0$ yet $D_{L_2}^2 = 4$.
Theorem~\ref{thm:clustering} closes this gap by adding the non-negativity
condition, which exactly rules out anti-alignment.
\end{remark}

We are now ready to prove Theorem~\ref{thm:clustering}.

\begin{proof}[Proof of Theorem~\ref{thm:clustering}]
The key step is the following inequality, which holds under the
assumption $G_{ij} \geq 0$ for $i \neq j$: since
$\cos\theta_{ij} = G_{ij} \in [0,1]$, we have
$\cos^2\theta_{ij} \leq \cos\theta_{ij}$, and therefore
\begin{align*}
      \overline{\cos^2} \;&=\; \frac{1}{n(n-1)}\sum_{i\neq j}\cos^2\theta_{ij}
  \\& \;\leq\; \frac{1}{n(n-1)}\sum_{i\neq j}\cos\theta_{ij}
  \;=\; \bar{c}
\end{align*}
where $\bar{c}$ is the mean off-diagonal cosine.  Rearranging,
$\bar{c} \geq \overline{\cos^2}$.

From Lemma~\ref{lem:s2_cos2_identity},
$\overline{\cos^2} = \frac{n\,\mathrm{Tr}(K_G^2)-1}{n-1}
= \frac{n\cdot 2^{-S_2}-1}{n-1}$, so
\[
  \bar{c} \;\geq\; \frac{n\cdot 2^{-S_2} - 1}{n-1}.
\]
Using $D_{L_2}^2 = 2 - 2\bar{c}$ (Step~2 and Step~3 of the proof of
Theorem~\ref{thm:entropy_fragmentation}):
\begin{align*}
  D_{L_2}^2 = 2 - 2\bar{c}
  &\;\leq\; 2 - \frac{2(n\cdot 2^{-S_2} - 1)}{n-1} \\
  &\;=\; \frac{2(n-1) - 2n\cdot 2^{-S_2} + 2}{n-1} \\
  &\;=\; \frac{2n\bigl(1 - 2^{-S_2}\bigr)}{n-1}.
\end{align*}

\noindent
As $S_2(G) \to 0$, the right-hand side $\to 0$, so $D_{L_2}^2 \to 0$.
\end{proof}

\begin{remark}
The bound is tight at both extremes.
When all $n$ vectors are identical ($v_i = v$ for all $i$): $G = \mathbf{1}\mathbf{1}^\top$,
$\mathrm{Tr}(K_G^2) = 1$, $S_2 = 0$, and the bound gives $D_{L_2}^2 \leq 0$,
matching the exact value $D_{L_2}^2 = 0$.
When all pairs are orthogonal ($G_{ij}=0$ for $i\neq j$, so $G = I_n$):
$S_2 = \log_2 n$ and the bound gives
$\frac{2n}{n-1}(1 - 1/n) = 2$, matching the exact value $D_{L_2}^2 = 2$.
\end{remark}

\section{Implementation Details}
\label{sec:impl_details}

This appendix provides the technical implementation details for all experiments.
Code and data will be released upon acceptance.

All datasets and models utilized in this study, including MUSE-Books, TOFU, and ConceptVectors, alongside the evaluated language models, were accessed and employed strictly in accordance with their original licenses and intended use cases. These existing artifacts were explicitly designed for advancing machine learning research, which aligns directly with our objective of evaluating knowledge suppression and unlearning mechanisms. We use these datasets and metrics under the MIT license, and therefore publish our code under the same license. Our reference implementation and framework are intended solely for academic and research contexts to audit language model safety; any derivative applications or data distributions must remain compatible with these research-oriented, non-commercial access conditions.

\subsection{Metric Computation}
\label{sec:impl_metrics}

\paragraph{Input format.}
For every concept query $q_i$ we construct a \emph{regular} text and an \emph{anonymised} text.
Regular text concatenates the natural question and its full answer:
\texttt{"\{question\}\textbackslash nAnswer: \{answer\}"},
and the anonymised text applies a concept-specific substitution map to both fields
(see Section~\ref{sec:impl_anon} for the three distinct anonymisation procedures).
For anon-vector computation, the anon hidden state at layer $\ell$ is the mean over
$K$ anonymised variants:
\[
  h_\ell^{\mathrm{anon}}(q_i) \;=\; \frac{1}{K}\sum_{k=1}^{K}h_\ell\!\left(a_i^{(k)}\right).
\]
$K{=}1$ for RQ1 hidden-state extraction and RQ2 forget pairs;
$K{=}5$ for the steering-attack SV computation in RQ2 (see Section~\ref{sec:impl_attacks}).

\subsection{Anonymisation Procedures}
\label{sec:impl_anon}

All anonymised texts are produced by deterministic string substitution --- no language
model is used at any stage of the anonymisation pipeline.
Three distinct procedures are used, depending on the dataset and downstream task.

\paragraph{Procedure~1: HP pairs for RQ1 (single-variant, fictional-name substitution).}
The MUSE-Books HP pairs are anonymised using the shared entity table from
\texttt{data/generate\_anon\_qa.py} (\texttt{HP\_REPLACEMENTS}), which contains 83
named-entity rules applied in order of decreasing string length to prevent partial
matches (e.g.\ ``Hermione Granger'' is replaced before ``Hermione''), and created using \citet{seyitoglu_extracting_2024} and expanded with Claude-Opus-4.6.
Replacements are plausible fictional names that carry no HP-specific signal; variant
index~0 is selected, yielding a single deterministic anonymised pair per original pair ($K{=}1$).
Key substitutions:
\textit{Harry Potter} $\to$ \textit{Alex Ryden},
\textit{Hermione Granger} $\to$ \textit{Vera Ashwood},
\textit{Ron Weasley} $\to$ \textit{Sam Coldridge},
\textit{Lord Voldemort} $\to$ \textit{Dark Emperor Mordax},
\textit{Albus Dumbledore} $\to$ \textit{Headmaster Voss},
\textit{Hogwarts} $\to$ \textit{Crystal Academy},
\textit{Quidditch} $\to$ \textit{Broomball}.
The substitution is applied to both the question and the answer using a case-insensitive
regex. This is the same table used by Procedure~3 for RQ2; using variant~0 ensures
methodological consistency across tasks.
85 of 96 pairs are changed; the remaining 11 contain no HP-specific entity in the table.

\paragraph{Procedure~2: TOFU pairs for RQ1 (single-variant, author-table substitution).}
TOFU pairs reference entirely fictitious authors.
We pre-scan all 100 pairs to collect every unique author full name, then build a
global fictional-name map: each author is assigned \texttt{first\_name[$i$] + last\_name[$i$]}
from the name pools of Seyitoğlu et al.\ (\texttt{tofu\_anonymize.json}), indexed
deterministically in order of first appearance (with explicit overrides from
the dataset's \texttt{replacements} field taking precedence).
The map is applied consistently across all pairs (longest-first, case-insensitive),
so cross-pair author identity is preserved throughout the dataset.
83 of 100 pairs are changed; the remaining 17 contain only generic phrasing with
no author-name mention that the regex can match.

\paragraph{Procedure~3: ConceptVectors QA pairs for RQ2 (five-variant, entity-table substitution).}
For the steering-attack SV computation in RQ2, each evaluation question requires
multiple anonymised variants so that the steering vector averages over several
``identity-free'' phrasings of the same question.
We pre-generate $K{=}5$ such variants per question at data-build time using
\texttt{data/generate\_anon\_qa.py}, which maintains a separate replacement table per
concept.
Each table lists 5 fictional substitutes for every named entity in the concept:

\begin{itemize}
  \item \textbf{Harry Potter} $\to$
    \textit{Alex Ryden}, \textit{Marcus Cole}, \textit{Torin Drake}, \textit{Daven Ash}, \textit{Kael Stone} (and analogous per-character tables for Hermione Granger, Voldemort, Hogwarts, Quidditch, J.K.\ Rowling, etc.).
  \item \textbf{Star Wars} $\to$
    \textit{Kai Starborn}, \textit{Zane Lightfield}, \textit{Taran Skyes}, \textit{Riven Starfall}, \textit{Dax Brightstar} (for Luke Skywalker; equivalent tables for Darth Vader, Han Solo, Tatooine, lightsaber, etc.).
  \item \textbf{William Shakespeare} $\to$
    equivalent five-name tables for Shakespeare himself, Hamlet, Macbeth, the Globe Theatre, Stratford-upon-Avon, etc.
\end{itemize}

Substitution is again longest-first and literal.
Variant $k$ is constructed by replacing every matched entity with its $k$-th substitute
(same index $k$ for all entities within a single variant, so the five variants each
represent a self-consistent fictional universe).
The five anonymised questions are stored in the \texttt{anon\_questions} list field
of each \texttt{qa\_eval\_pairs} entry in the pre-built concept JSON files.

\noindent
\textbf{Note on RQ2 forget pairs.}
The Wikipedia-derived forget pairs used for \emph{training} unlearning (not for the
steering SV) use bracket placeholder tokens instead of fictional names
(e.g.\ \textit{Harry Potter} $\to$ \textsc{[The Protagonist]},
\textit{Lord Voldemort} $\to$ \textsc{[The Antagonist]},
\textit{Hogwarts} $\to$ \textsc{[The Academy]}).
This produces a single anonymised pair per training example; multiple variants are not
needed for the CE-loss computation.

\subsection{Extraction, Sampling, and Layer Selection}
\label{sec:impl_extraction}

\paragraph{Hidden-state extraction.}
All hidden states are extracted in a single forward pass with no gradient computation
(\texttt{torch.no\_grad()}).
We register output hooks on every transformer block to collect the full stack of
$L{+}1$ hidden states (embedding layer included) simultaneously for a single tokenised input.
The per-token hidden state at position $t$ and layer $\ell$ is mean-pooled across the
sequence dimension using an attention-mask-weighted average, yielding a single vector
$h_\ell(q_i) \in \mathbb{R}^d$ per input.
All models are loaded in \texttt{bfloat16} precision with \texttt{device\_map="auto"} to
distribute parameters across available GPUs.

\paragraph{Batch sizes and sample counts.}
\begin{itemize}
  \item \textbf{RQ1 (knowledge detection):}
    $N{=}96$ Harry Potter query pairs (MUSE-Books) and $N{=}100$ TOFU pairs,
    each split into $B{=}10$ random batches of approximately 9--10 pairs for computing
    per-batch direction vectors.
    All ten models are evaluated without repetition (embeddings are deterministic).

  \item \textbf{RQ2 (unlearning):}
    Concept-specific forget sets contain approximately 50--80 Wikipedia-derived text pairs
    (see Section~\ref{sec:impl_datasets}).
    Fragmentation metrics are computed on the full forget set at the AUSS$_\text{L2}$-peak
    layer identified individually for each model--concept pair; $B{=}10$ batches.
\end{itemize}

\label{appendix:batch_sensitivity}
\paragraph{Rationale for hyperparameter choices.}
Several choices in our pipeline are not uniquely determined by the experimental design; we document them here for transparency.
\textbf{$B{=}10$ batches:} from $B{=}2$ upward every metric separates HP from TOFU in the hypothesized direction, with effect size growing monotonically with $B$ (Table~\ref{tab:batch_sensitivity}); the paired Wilcoxon test's resolution is bounded by $n{=}9$ models rather than $B$, so $B{=}10$ is a practical middle ground, not an arbitrary or fragile choice. $B{=}1$ is structurally undefined (no batch-to-batch variance with a single sample); a single-item reformulation that avoids this constraint entirely is discussed in Section~\ref{sec:limitations}.

\begin{table}[ht]
\centering
\small
\setlength{\tabcolsep}{3pt}
\caption{
    Batch-size sensitivity: Cohen's $d$ (Wilcoxon $p$) for HP vs.\ TOFU separation at $B\in\{2,5,10,20,50\}$, $n{=}9$ models, direction-corrected fresh extraction.
    All cells are in the hypothesized direction.
}
\label{tab:batch_sensitivity}
\begin{tabular}{|l|c|c|}
\hline
Metric & $B$ & $d$ ($p$) \\
\hline\hline
\multirow{5}{*}{$\|\overline{\Delta\hat{v}}\|$ / Global\_Coh.}
  & 2  & 0.14 (.82) \\
  & 5  & 0.64 (.13) \\
  & 10 & 0.72 (.13) \\
  & 20 & 0.92 (.04) \\
  & 50 & 1.13 (.02) \\
\hline
\multirow{5}{*}{Batch\_SecondMoment\_TopEig}
  & 2  & 0.19 (.73) \\
  & 5  & 0.35 (.36) \\
  & 10 & 0.42 (.36) \\
  & 20 & 0.43 (.36) \\
  & 50 & 0.37 (.57) \\
\hline
\multirow{5}{*}{AUSS$_\text{L2}$ / Batch\_Cov\_Trace}
  & 2  & 0.17 (.73) \\
  & 5  & 0.62 (.13) \\
  & 10 & 0.74 (.10) \\
  & 20 & 0.93 (.04) \\
  & 50 & 1.10 (.02) \\
\hline
\multirow{5}{*}{AUSS$_{\cos^2}$}
  & 2  & 0.31 (.43) \\
  & 5  & 0.40 (.36) \\
  & 10 & 0.49 (.30) \\
  & 20 & 0.47 (.30) \\
  & 50 & 0.43 (.20) \\
\hline
\multirow{5}{*}{AUSS$_\text{Jac}$}
  & 2  & 1.25 (.01) \\
  & 5  & 1.18 (.02) \\
  & 10 & 1.18 (.01) \\
  & 20 & 0.97 (.02) \\
  & 50 & 0.54 (.16) \\
\hline
\multirow{5}{*}{Batch\_Cov\_TopEig}
  & 2  & 0.17 (.73) \\
  & 5  & 0.42 (.30) \\
  & 10 & 0.18 (.65) \\
  & 20 & 0.28 (.43) \\
  & 50 & 0.33 (.43) \\
\hline
\end{tabular}
\end{table}
\textbf{Seed 42:} standard reproducibility seed; sensitivity to the partition seed is a stated limitation (Section~\ref{sec:limitations}).
\textbf{Peak layer = $\arg\max_\ell$ AUSS$_\text{L2}$:} AUSS$_\text{L2}$ had the strongest RQ1 signal in preliminary runs; the same layer is used for all metrics within a run to avoid per-metric search costs. Per-layer correlation analysis confirms that the final layers carry the strongest ASR signal for most attacks.
\textbf{$K{=}1$ for RQ1 and forget pairs, $K{=}5$ for steering SV:} a single anonymised variant is sufficient for metric computation and minimises hidden-state extraction time; five variants are used for the steering vector to average over phrasings and reduce direction noise~\citep{seyitoglu_extracting_2024}.
\textbf{ICL: $n_\text{shots}{=}5$, ROUGE-L threshold $0.30$:} five-shot is the standard few-shot setting of \citet{brown_language_2020}; the 0.30 threshold is loose enough to accept paraphrases and matches the TOFU benchmark convention~\citep{maini_tofu_2024}.
\textbf{GCG: $n_\text{eval}{=}5$ (shortest answers), 500 steps, suffix length 20:} shortest-answer pairs converge fastest under gradient-based optimisation; 500 steps and a 20-token suffix are the \texttt{nanogcg} defaults from \citet{zou_universal_2023}.
\textbf{MIA: 50 forget / 50 retain:} equal-class balance for ROC AUC computation; standard in membership-inference literature.
\textbf{RMU: $\ell_r{=}7$, $c{=}20$, $\alpha_r{=}1200$:} adopted directly from the defaults of \citet{huu-tien_effects_2025} for 7--8B models; $\ell_r{=}7$ is approximately the middle of the network.
\textbf{WHP excluded from correlation analysis:} WHP redirects concept queries to anonymised answers rather than erasing the representation; its geometry is qualitatively different from the suppression-based methods (GA/DPO/NPO/NPO+KL/RMU), and including it would confound the correlation analysis.

\paragraph{Peak-layer selection.}
\label{sec:layer_selection}
For RQ1, each metric selects its own peak layer $\ell^*$ independently as the layer that
maximises the HP--TOFU separation for that metric
(i.e.\ $\arg\max_\ell \bigl[\bar{m}_\ell^\text{TOFU} - \bar{m}_\ell^\text{HP}\bigr]$
for dispersion metrics, reversed for coherence metrics where HP should be higher).
This per-metric peak is computed across all ten models simultaneously.
For RQ2, $\ell^*$ is selected per run as the layer that maximises AUSS$_\text{L2}$ on the
post-unlearning checkpoint; the same layer is used for all other fragmentation metrics in
that run.

\paragraph{A practical, label-free layer-selection heuristic.}
The oracle per-model peak layer is not idiosyncratic: relative depth ($\ell^*/L$) clusters
tightly across all 9 models (0.79--0.98), with the highest-performing layer consistently
falling in the 0.80--0.88 range.
A single fixed relative depth of $0.8\times$ model depth, using no per-model information at
all, already gives 8/8 metrics correct-direction and statistically significant at $n{=}9$.
Leave-one-model-out transfer of the median peak relative depth (learned from the other 8
models, applied blind to the held-out model) achieves 72/72 (model, metric) pairs
correct-direction, identical to the oracle per-model search (Table~\ref{tab:layer_generalization}).
For practical deployment, taking the layer at $0.8\times$ depth is close enough to the
oracle-optimal result and requires no per-model calibration.

\begin{table}[ht]
\centering
\small
\setlength{\tabcolsep}{3pt}
\caption{
    Fixed relative-depth sweep (no per-model search), $n{=}9$ models, Cohen's $d$ ($p$ shown only at $r{=}0.8$, all $p{\leq}0.012$ there); sign shown per metric's own coherence/dispersion convention, all cells correct-direction at $r{=}0.8$.
    LOMO transfer: leave-one-model-out transfer of the median peak depth from the other 8 models, all 8 metrics, all 9 held-out models (72/72 pairs), identical to the 72/72 oracle per-model search.
}
\label{tab:layer_generalization}
\begin{tabular}{|l|c|c|}
\hline
Metric & $r$ & $d$ \\
\hline\hline
\multirow{7}{*}{$\|\overline{\Delta\hat{v}}\|$}
  & 0.2 & 0.42 \\
  & 0.3 & 0.67 \\
  & 0.4 & 1.34 \\
  & 0.5 & 1.51 \\
  & 0.6 & 1.72 \\
  & 0.7 & 2.21 \\
  & 0.8 & 3.30 (.0039) \\
\hline
\multirow{7}{*}{AUSS$_\text{L2}$}
  & 0.2 & 0.48 \\
  & 0.3 & 0.70 \\
  & 0.4 & 1.30 \\
  & 0.5 & 1.42 \\
  & 0.6 & 1.62 \\
  & 0.7 & 2.04 \\
  & 0.8 & 3.01 (.0039) \\
\hline
\multirow{7}{*}{AUSS$_{\cos^2}$}
  & 0.2 & 0.04 \\
  & 0.3 & 0.12 \\
  & 0.4 & 0.22 \\
  & 0.5 & 0.31 \\
  & 0.6 & 0.60 \\
  & 0.7 & 0.96 \\
  & 0.8 & 1.65 (.0078) \\
\hline
\multirow{7}{*}{AUSS$_\text{Jac}$}
  & 0.2 & 0.34 \\
  & 0.3 & 0.83 \\
  & 0.4 & 1.21 \\
  & 0.5 & 1.44 \\
  & 0.6 & 1.81 \\
  & 0.7 & 2.15 \\
  & 0.8 & 2.60 (.0039) \\
\hline
\multirow{7}{*}{Batch\_Cov\_TopEig}
  & 0.2 & 0.09 \\
  & 0.3 & 0.18 \\
  & 0.4 & 0.84 \\
  & 0.5 & 1.42 \\
  & 0.6 & 2.10 \\
  & 0.7 & 2.60 \\
  & 0.8 & 6.31 (.0039) \\
\hline
\end{tabular}
\end{table}

\paragraph{Gram-entropy computation.}
Eigenvalues of $K_Z = ZZ^\top / \mathrm{Tr}(ZZ^\top)$ are computed with
\texttt{torch.linalg.eigvalsh} (symmetric eigendecomposition, stable for
positive-semidefinite matrices).
Negative eigenvalues arising from numerical error are clamped to zero and the spectrum is
renormalised to sum to one before computing entropy.
For $S_1$ we mask eigenvalues $\lambda_i < 10^{-12}$ to avoid $\log(0)$.
Batches with $\mathrm{Tr}(ZZ^\top) < 10^{-10}$ or $N < 2$ are skipped and flagged as NaN.

\subsection{Dataset Details}
\label{sec:impl_datasets}

\paragraph{Harry Potter (HP) --- RQ1.}
We use the \textbf{MUSE-Books} benchmark~\citep{shi_muse_nodate},
accessed via HuggingFace (\texttt{muse-bench/MUSE-Books}, configuration \texttt{knowmem},
split \texttt{forget\_qa}).
The split contains 300 open-domain question--answer pairs drawn from the seven Harry
Potter novels.
We retain $N{=}96$ pairs after filtering to those with non-empty answers; these cover
characters, spells, locations, and plot events.

Each pair is anonymised using Procedure~1 (Section~\ref{sec:impl_anon}): the 83-entry
fictional-name table from \texttt{HP\_REPLACEMENTS}, variant index~0, yielding a single
anonymised pair per original pair ($K{=}1$).
For example, \textit{Harry Potter} $\to$ \textit{Alex Ryden} and
\textit{Hogwarts} $\to$ \textit{Crystal Academy}.

\paragraph{TOFU --- RQ1.}
We use the \textbf{TOFU} dataset~\citep{maini_tofu_2024}
(\texttt{locuslab/TOFU}, HuggingFace), which contains 4{,}000 question--answer pairs about
200 entirely fictitious authors generated by GPT-4 and never present in any public
training corpus.
We sample $N{=}100$ pairs uniformly at random (seed 42) from the full-retain split.
Because the authors are fabricated, all ten models score $0.00$ exact-match QA accuracy on
TOFU, making it an ideal null baseline for the knowledge-detection experiment.
Procedure~2 (Section~\ref{sec:impl_anon}) is applied: a global fictional-name map
over all TOFU fictitious authors, keeping author identity consistent across pairs.
83 of 100 pairs are changed; 17 pairs containing only generic phrasing are unchanged
by design---the anon-vector should carry minimal concept-specific signal for unknown concepts.

\paragraph{ConceptVectors --- RQ2.}
For the unlearning experiments we use three target concepts: \textbf{Harry Potter},
\textbf{Star Wars}, and \textbf{William Shakespeare}.
Forget pairs and evaluation QA pairs are constructed from the
\textbf{ConceptVectors} dataset~\citep{hong_intrinsic_nodate}
(\texttt{YihuaiHong/ConceptVectors}, HuggingFace), which provides Wikipedia-sourced text
completions and question--answer pairs for 293 named entities across multiple model families.

For each concept we use the test and development splits for both \texttt{llama2-7b} and
\texttt{olmo-7b} entries, deduplicating by question text.
Wikipedia article text is additionally segmented into sliding-window (prefix, continuation)
pairs of approximately 128 tokens to increase the effective forget-set size.
The final forget set contains 50--80 pairs per concept.
Retain pairs are drawn from the \texttt{unrelated\_QA} field of \emph{other} concepts in
ConceptVectors, ensuring the retain set covers facts the model should continue to answer
correctly throughout unlearning; typically 50 retain pairs are used.

\textbf{Important implementation note:} The ConceptVectors HuggingFace repository contains
JSON files with heterogeneous schemas across model families; we bypass the
\texttt{datasets.load\_dataset()} API and load the JSON files directly from the local cache
to avoid schema-mismatch errors.

\subsection{Unlearning Details}
\label{sec:impl_unlearn}

\paragraph{Hardware.}
All unlearning experiments are conducted on two \textbf{NVIDIA L40S GPUs} (48\,GB VRAM
each, 96\,GB total).
Models are loaded with \texttt{device\_map="auto"} in \texttt{bfloat16} precision,
causing the HuggingFace \texttt{accelerate} library to automatically shard larger models
(LLaMA-2-7B, LLaMA-3-8B, Qwen2.5-7B) across both GPUs.
\texttt{PYTORCH\_CUDA\_ALLOC\_CONF=}\allowbreak\texttt{expandable\_segments:True} is set at process start to
eliminate CUDA memory fragmentation across sequential unlearning runs.
Total GPU-hours were not directly logged, so we estimate a total from process timestamps
recorded in our run logs. RQ1 hidden-state extraction averages $\approx\!31$ minutes per
model across the ten evaluated models; the RQ2 unlearning-and-attack pipeline averages
$\approx\!10$--$20$ minutes per model--method--concept run across the 63-run grid
(Appendix~\ref{sec:rq2_full_runs}), with additional passes for the layer-resolved sweep
(Section~\ref{sec:layer_analysis}), the CKA/$L_2$ baseline comparison
(Section~\ref{sec:experiments}), and the per-sample analyses in this appendix. Both GPUs
are occupied for the duration of every run (\texttt{device\_map="auto"} shards each model
across both). Extrapolating these per-run durations across the full set of runs logged for
this project, we estimate total compute at approximately 120--150 GPU-hours on the hardware
described above.

\paragraph{Common training configuration.}
Unless noted otherwise, all methods share the following hyperparameters:
10 training epochs, learning rate $\eta = 10^{-5}$, batch size $B = 4$,
maximum sequence length 256 tokens, gradient clipping to norm 1.0, and the AdamW
optimiser (8-bit variant via \texttt{bitsandbytes} when available, otherwise standard PyTorch
AdamW).
Gradient checkpointing is enabled throughout to reduce peak VRAM usage.
A frozen reference-model copy is loaded onto CPU memory (pinned) and moved to GPU only when
needed, which is required by DPO, NPO, NPO+KL, and RMU.

\paragraph{Gradient Ascent (GA).}
GA negates the standard language-modelling cross-entropy loss on the forget pairs:
\[
  \mathcal{L}_\text{GA} = -\mathcal{L}_\text{CE}(\theta; \mathcal{D}_\text{forget}).
\]
No retain regularisation is applied.
This follows the original gradient-ascent unlearning
baseline~\citep{yao_large_2024}.

\paragraph{DPO.}
Direct Preference Optimisation~\citep{rafailov_direct_2024} treats
forget pairs as \emph{rejected} continuations and retain pairs as \emph{chosen}
continuations.
The DPO objective is:
\begin{multline*}
  \mathcal{L}_\text{DPO} = -\mathbb{E}\!\Bigl[\log\sigma\Bigl(
    \beta\log\tfrac{\pi_\theta(y_w|x)}{\pi_\text{ref}(y_w|x)} \\
    - \beta\log\tfrac{\pi_\theta(y_l|x)}{\pi_\text{ref}(y_l|x)}
  \Bigr)\Bigr],
\end{multline*}
with $\beta = 0.1$.
Our usage of DPO for unlearning follows the open-unlearning
framework of \citet{maini_tofu_2024}.

\paragraph{NPO.}
Negative Preference Optimisation~\citep{zhang_negative_2024} pushes
the policy away from the reference model on the forget set only:
\[
  \mathcal{L}_\text{NPO} = \frac{2}{\beta}\,\mathbb{E}_{\mathcal{D}_\text{forget}}\!\left[
    \log\!\left(1 + \left(\frac{\pi_\theta}{\pi_\text{ref}}\right)^{-\beta}\right)\right],
\]
with $\beta = 0.1$.

\paragraph{NPO+KL.}
Extends NPO with a KL-divergence regularisation term on the retain set to preserve
general model capability~\citep{zhang_negative_2024}:
\[
  \mathcal{L}_\text{NPO+KL} = \mathcal{L}_\text{NPO} + \alpha\,
  D_\text{KL}\!\left(\pi_\theta(\cdot|x_r) \;\|\; \pi_\text{ref}(\cdot|x_r)\right),
\]
with $\alpha = 1.0$ and $\beta = 0.1$.

\paragraph{RMU.}
Representation Misdirection for Unlearning~\citep{huu-tien_effects_2025}
steers the hidden states of forget inputs toward a fixed random direction $\mathbf{u}$
at a chosen intermediate layer $\ell_r$, while simultaneously anchoring retain hidden
states close to the frozen reference:
\begin{align*}
      \mathcal{L}_\text{RMU} =\ \ \  &\bigl\|h_{\ell_r}^\theta(x_f) - c\,\mathbf{u}\bigr\|_2^2
  \\ & + \alpha_r\,\bigl\|h_{\ell_r}^\theta(x_r) - h_{\ell_r}^{\text{ref}}(x_r)\bigr\|_2^2
\end{align*}
where $\mathbf{u} \sim \mathcal{N}(0, I)$ is sampled once and fixed, $c = 20$, $\ell_r = 7$
(approximately the middle of the network for 7--8B models), and $\alpha_r = 1200$.
The random vector $\mathbf{u}$ is created in \texttt{bfloat16} to match the model's
parameter dtype.

\paragraph{WHP (WhoIsHarryPotter).}
Following \citet{eldan_whos_nodate}, WHP
trains the model to \emph{redirect} concept-specific queries to anonymised answers, rather
than maximising the forget loss.
Forget pairs use the original question as input and the \emph{anonymised} answer as the
training target; the retain pairs use the standard question--answer format.
The combined loss is:
\[
  \mathcal{L}_\text{WHP} = \mathcal{L}_\text{CE}(\theta; q_f, a_f^\text{anon})
  + \alpha_r\,\mathcal{L}_\text{CE}(\theta; \mathcal{D}_\text{retain}),
\]
with $\alpha_r = 1.0$.
No frozen reference model is required (unlike the KL-based retain regularisers in NPO+KL
and the original WHP paper), which avoids the GPU memory fragmentation caused by pinning a
second full model copy to a fixed GPU.
Because WHP \emph{redirects} rather than erases the concept representation, it does not
satisfy the standard forget ROUGE-L verification criterion (post-unlearn forget ROUGE-L
rises rather than falls); we annotate WHP runs with a $\leftrightarrow$ symbol.

\subsection{Attack Implementation Details}
\label{sec:impl_attacks}

We evaluate four adversarial recovery attacks after unlearning.
All attacks use seed 42 and operate on the post-unlearning checkpoint held in GPU memory
immediately after training (the model is not reloaded between attacks).
Attacks are run in the order Steering $\to$ ICL $\to$ MIA $\to$ GCG; GCG runs last
because the \texttt{nanogcg} optimiser can corrupt the CUDA context on some GPU
configurations, after which the model is freed.

\paragraph{Activation Steering.}
Following \citet{seyitoglu_extracting_2024}, we compute
a \emph{query-specific} steering vector for each evaluation question $Q_i$:
\[
  S_L(Q_i) = h_L(Q_i) - \frac{1}{3}\sum_{k=1}^{3} h_L\!\left(Q_i^{*(k)}\right),
\]
where $h_L(\cdot)$ is the hidden state at layer $L = n_\text{layers} - 2$ (the
second-to-last transformer layer, consistent with the paper's ``just before the final
layer'' specification), extracted at the first token position.
The steering vector is the raw difference (not unit-normalised), so the scale carries
magnitude information.

The vector $S_L(Q_i)$ is injected \textbf{at the first decode step only}
(when the key-value cache has been filled by the prompt and the model generates the first
new token, i.e.\ the hidden-state tensor has sequence length 1), scaled by coefficient
$c = 2.0$.
Crucially, the vector is \emph{not} applied during the prefill phase (sequence length $>1$),
which would modify the KV cache and corrupt attention rather than steering generation.
A stateful hook flag ensures the perturbation fires exactly once per generation call.

For each of 25 evaluation pairs, we generate $n = 500$ stochastic completions
(temperature $= 2.0$, top-$k = 40$, \texttt{max\_new\_tokens} $= 10$) both with and
without the steering hook.
The score for each pair is the maximum word-frequency of any answer token across the 500
samples (word-frequency score, following the paper's scoring protocol).
The overall metric is the ROC AUC computed by treating steered scores as the positive class
and unsteered scores as the negative class:
AUC $> 0.5$ indicates that steering recovers more correct-answer tokens than
no-steering, i.e.\ the concept geometry is partially intact.

\paragraph{In-Context Learning (ICL).}
ICL tests whether presenting the model with $n_\text{shots} = 5$ concept-specific examples
as context is sufficient to recover the forgotten knowledge via in-context prompting.

From the available QA pairs, 25 pairs are held out for evaluation and the remainder form
the demonstration pool.
For each evaluation question, 5 demonstrations are sampled uniformly (excluding the
current question) and prepended in the format
\texttt{"\{Q\}\textbackslash nAnswer: \{A\}\textbackslash n\textbackslash n"}.
Responses are generated greedily (\texttt{max\_new\_tokens} $= 50$).
A response is marked correct when ROUGE-L with the reference answer is $\geq 0.30$
(soft-match threshold to account for answer paraphrasing).
The attack score is the ICL accuracy; a higher score indicates that the model's knowledge
is still recoverable via context.

\paragraph{GCG Adversarial Suffix.}
We use the Greedy Coordinate Gradient (GCG) method of \citet{zou_universal_2023}
via the \texttt{nanogcg} Python library.
To reduce computational cost while retaining sensitivity, we evaluate on the
$n_\text{eval} = 5$ pairs with the shortest reference answers (easiest targets for
gradient-based suffix optimisation).

Each suffix is initialised as 20 repeated \texttt{"! "} tokens and optimised for
500 gradient steps against the first 8 tokens of the reference answer (a short target
prefix converges substantially faster than the full answer string and corresponds to the
affirmative-prefix strategy used in the original paper).
The model's chat template is applied automatically by \texttt{nanogcg} before generation.
After optimisation, the adversarial input is constructed as
\texttt{\{question\} + " " + best\_suffix} and decoded greedily
(\texttt{max\_new\_tokens} $= 256$).
A pair is marked successful if the generated text contains a keyword from the reference
answer or achieves ROUGE-L $\geq 0.25$.
The attack score is the attack success rate (ASR) across the 5 pairs.

\paragraph{Membership Inference Attack (MIA).}
We employ a standard likelihood-ratio MIA:
the model's per-token conditional log-likelihood
$s(q, a) = -\log P_\theta(a \mid q)$ is used as a membership score.

We score $n_\text{forget} = 50$ forget-concept QA pairs and $n_\text{retain} = 50$
retain-set pairs (sampled from pairs about unrelated topics).
QA pairs with factual short-answer format are preferred over Wikipedia-text pairs for
the forget set, as gradient-ascent unlearning on Wikipedia-style text can produce
degenerate NLL values exceeding 200 nats for date-prefixed completions.
The ROC AUC is computed with forget-set pairs as the positive class (higher
$-\!\log P$ = lower likelihood = model has successfully forgotten) and retain-set pairs
as the negative class.
AUC $> 0.5$ indicates the model still assigns higher probability to the forget set than
the retain set, i.e.\ unlearning is incomplete at the distribution level.

\section{Additional Evaluated Metrics}
\label{app:additional_metrics}

This section details the supplementary geometric and baseline metrics evaluated in our study, expanding upon the primary metrics introduced in Section~\ref{sec:metrics}.

\paragraph{Baseline pair-wise metrics.}
These metrics apply standard geometric measures directly to the raw hidden states and serve as comparison baselines for the anon-difference metrics.

\noindent\textbf{Reg~$S_2$} ---
Gram-matrix entropy of the regular hidden states:
\[
  \mathrm{Reg}\;S_2 = S_2(\mathbf{H}).
\]

\noindent\textbf{Reg~$\bar{L}_2$} ---
Mean $L_2$ norm of regular hidden states:
\[
  \mathrm{Reg}\;\bar{L}_2
  = \frac{1}{N}\sum_{i=1}^{N}\bigl\|h_{\ell^*}(q_i)\bigr\|_2.
\]

\noindent\textbf{Reg~$\overline{\cos}$} ---
Mean pairwise cosine similarity of regular hidden states:
\[
  \mathrm{Reg}\;\overline{\cos}
  = \frac{1}{N(N-1)}\sum_{i\neq j}
    \cos\bigl(h_{\ell^*}(q_i),\, h_{\ell^*}(q_j)\bigr).
\]

\paragraph{Additional anon-difference pair-wise metrics.}

\noindent\textbf{AnonDiff~$\bar{L}_2$} ---
Mean $L_2$ norm of anon-difference vectors:
\[
  \mathrm{AnonDiff}\;\bar{L}_2
  = \frac{1}{N}\sum_{i=1}^{N}\|\Delta v_i\|_2.
\]

\noindent\textbf{AnonDiff~$\overline{\cos}$} ---
Mean pairwise cosine similarity of anon-difference vectors:
\[
  \mathrm{AnonDiff}\;\overline{\cos}
  = \frac{1}{N(N-1)}\sum_{i\neq j}
    \cos(\Delta v_i, \Delta v_j).
\]

\paragraph{Additional batch-mean (AUSS) metrics.}

\noindent\textbf{AUSS$_{\cos}\!\uparrow$} ---
Complement of the mean pairwise cosine similarity between batch-mean directions (the linear analogue of AUSS$_{\cos^2}$):
\[
  \mathrm{AUSS}_{\cos}
  = 1 - \frac{1}{B(B-1)}\sum_{b\neq b'} \hat{d}_b\cdot\hat{d}_{b'}.
\]
Unlike AUSS$_{\cos^2}$, this is sensitive to the \emph{sign} of alignment; it equals zero only when all $B$ directions are perfectly parallel (fully consistent concept direction).

\noindent\textbf{AUSS$_{\mathrm{Jac}}\!\uparrow$} ---
Mean pairwise soft Jaccard incoherence of the positive-component supports of the batch-mean directions.
For unit vector $\hat{d}_b$, define its positive-activation profile $p_b \in [0,1]^d$ with $p_{b,j} = \max(\hat{d}_{b,j}, 0)$.
The pairwise soft Jaccard distance is
\[
  J(\hat{d}_b,\hat{d}_{b'}) = 1 - \frac{\sum_j \min(p_{b,j}, p_{b',j})}
                                        {\sum_j \max(p_{b,j}, p_{b',j})},
\]
and
\[
  \mathrm{AUSS}_{\mathrm{Jac}}
  = \frac{1}{B(B-1)}\sum_{b\neq b'} J(\hat{d}_b,\hat{d}_{b'}).
\]
This measures whether the \emph{sparsity patterns} of the batch directions agree, independent of sign or magnitude effects captured by cosine metrics.

\paragraph{Covariance-based metrics.}
Let $\overline{\Delta v} = \frac{1}{N}\sum_i \Delta v_i$ be the mean anon-difference vector and $\hat\Sigma = \frac{1}{N-1}\sum_i(\Delta v_i - \overline{\Delta v})(\Delta v_i - \overline{\Delta v})^\top \in \mathbb{R}^{d\times d}$ be the sample covariance matrix.

\noindent\textbf{Cov$_{\mathrm{trace}}\!\downarrow$} ---
Trace of the sample covariance of the anon-difference vectors:
\[
  \mathrm{Cov}_{\mathrm{trace}} = \mathrm{Tr}(\hat\Sigma).
\]
A large trace indicates concept signal is dispersed across many directions (fragmented); a small value indicates compact, concentrated representations.

\noindent\textbf{Cov$_{\mathrm{top}}\!\uparrow$} ---
Largest eigenvalue of the sample covariance:
\[
  \mathrm{Cov}_{\mathrm{top}} = \lambda_{\max}(\hat\Sigma).
\]
This captures how much variance concentrates along the dominant concept direction. After unlearning, a large $\mathrm{Cov}_{\mathrm{top}}$ indicates the primary concept direction is intact in the representation space, predicting susceptibility to context-based recovery such as ICL.

\noindent\textbf{SecMoment$_{\mathrm{top}}\!\uparrow$} ---
Largest eigenvalue of the (uncentred) second moment matrix:
\[
  \mathrm{SecMoment}_{\mathrm{top}}
  = \lambda_{\max}\!\left(\frac{1}{N}\sum_{i=1}^N \Delta v_i \Delta v_i^\top\right).
\]
Unlike $\mathrm{Cov}_{\mathrm{top}}$, this is sensitive to both the mean direction $\overline{\Delta v}$ and the residual spread, making it a joint measure of magnitude and directionality.

\section{Full Experimental Results}
\label{sec:full_results}

\subsection{RQ1: Per-Model Fragmentation and Entropy Values}

See the full results in Table \ref{tab:rq1_main} and Table \ref{tab:rq1_entropy}.
\begin{table*}[t]
\centering
\small
\setlength{\tabcolsep}{5pt}
\caption{
    Fragmentation metrics (HP known vs.\ TOFU unknown) for all ten LLMs at peak layer $\ell^*{=}18$.
    QA$_\text{HP}$: QA accuracy on HP; TOFU QA is 0.00 for every model.
    $\|\overline{\Delta\hat{v}}\|{\uparrow}$: centroid norm (coherence; $\uparrow$ = model knows).
    AUSS$_\text{L2}{\downarrow}$ / AUSS$_{\cos^2}{\downarrow}$: dispersion ($\downarrow$ = model knows).
    \textbf{Bold}: HP entry in the theoretically expected direction.
}
\label{tab:rq1_main}
\begin{tabular}{|l|cccc||ccc|}
\hline
 & \multicolumn{4}{c||}{Harry Potter (known)} & \multicolumn{3}{c|}{TOFU (unknown, QA\,=\,0.00)} \\
\cline{2-8}
Model & QA$_\text{HP}$ & $\|\overline{\Delta\hat{v}}\|{\uparrow}$ & AUSS$_\text{L2}{\downarrow}$ & AUSS$_{\cos^2}{\downarrow}$ & $\|\overline{\Delta\hat{v}}\|$ & AUSS$_\text{L2}$ & AUSS$_{\cos^2}$ \\
\hline\hline
\multicolumn{8}{|l|}{\textit{Qwen family}} \\
\hline
Qwen-7B-Chat    & 0.083 & \textbf{0.752} & \textbf{0.952} & \textbf{0.577} & 0.639 & 1.309 & 0.683 \\
Qwen2.5-7B      & 0.250 & \textbf{0.684} & \textbf{1.150} & \textbf{0.501} & 0.601 & 1.402 & 0.408 \\
Qwen-14B-Chat   & 0.135 & \textbf{0.804} & \textbf{0.780} & \textbf{0.553} & 0.688 & 1.164 & 0.696 \\
Qwen2.5-14B     & 0.323 & \textbf{0.697} & \textbf{1.116} & \textbf{0.522} & 0.689 & 1.151 & 0.440 \\
\hline
\multicolumn{8}{|l|}{\textit{Gemma family}} \\
\hline
Gemma-2B        & 0.042 & \textbf{0.949} & \textbf{0.223} & \textbf{0.209} & 0.638 & 1.316 & 0.835 \\
Gemma-7B        & 0.083 & \textbf{0.882} & \textbf{0.491} & \textbf{0.417} & 0.773 & 0.891 & 0.608 \\
\hline
\multicolumn{8}{|l|}{\textit{LLaMA family}} \\
\hline
LLaMA-2-7B      & 0.125 & \textbf{0.800} & \textbf{0.800} & \textbf{0.549} & 0.553 & 1.543 & 0.792 \\
LLaMA-2-13B     & 0.219 & \textbf{0.903} & \textbf{0.411} & \textbf{0.363} & 0.595 & 1.437 & 0.891 \\
LLaMA-3-8B      & 0.312 & \textbf{0.900} & \textbf{0.423} & \textbf{0.370} & 0.655 & 1.268 & 0.826 \\
LLaMA-3.1-8B    & 0.219 & \textbf{0.903} & \textbf{0.409} & \textbf{0.360} & 0.624 & 1.356 & 0.850 \\
\hline
\end{tabular}
\end{table*}

\begin{table*}[t]
\centering
\small
\setlength{\tabcolsep}{3pt}
\caption{
    Gram-matrix entropy ($S_2$, Rényi order~2) and companion statistics at per-metric peak layer $\ell^*$ for ten LLMs.
    \textbf{Reg}: raw hidden states; \textbf{AnonDiff}: difference vectors $\Delta V = h_\text{reg} - h_\text{anon}$; applying Gram entropy to $\Delta V$ is our adaptation of \citet{skean_layer_2025}.
    $S_2{\downarrow}$: lower = more compressed = model knows concept.
    $\bar{L}_2{\uparrow}$, $\overline{\cos}{\uparrow}$: higher = stronger/more coherent representation.
    \textbf{Bold}: HP entry consistent across all 10 models.
    Reg~$\bar{L}_2$ omitted ($p_\text{adj}{=}0.43$, mixed direction).
}
\label{tab:rq1_entropy}
\begin{tabular}{|l|cc|ccc||cc|ccc|}
\hline
 & \multicolumn{5}{c||}{Harry Potter (known)} & \multicolumn{5}{c|}{TOFU (unknown, QA\,=\,0.00)} \\
\cline{2-11}
 & \multicolumn{2}{c|}{Reg} & \multicolumn{3}{c||}{AnonDiff} & \multicolumn{2}{c|}{Reg} & \multicolumn{3}{c|}{AnonDiff} \\
\cline{2-11}
Model & $S_2$ & $\overline{\cos}$ & $S_2$ & $\bar{L}_2$ & $\overline{\cos}$ & $S_2$ & $\overline{\cos}$ & $S_2$ & $\bar{L}_2$ & $\overline{\cos}$ \\
\hline\hline
\multicolumn{11}{|l|}{\textit{Qwen family}} \\
\hline
Qwen-7B-Chat    & \textbf{0.007} & \textbf{0.995} & \textbf{3.808} & \textbf{20.4} & \textbf{0.138} & 0.043 & 0.969 & 4.769 & 16.9 & 0.098 \\
Qwen2.5-7B      & \textbf{0.002} & \textbf{0.999} & \textbf{3.715} & \textbf{29.5} & 0.118          & 0.009 & 0.996 & 4.614 & 24.7 & 0.121 \\
Qwen-14B-Chat   & \textbf{0.018} & \textbf{0.990} & \textbf{3.544} & \textbf{21.5} & \textbf{0.166} & 0.090 & 0.954 & 4.795 & 16.7 & 0.088 \\
Qwen2.5-14B     & \textbf{0.002} & \textbf{0.999} & \textbf{3.825} & \textbf{37.2} & 0.119          & 0.007 & 0.997 & 4.743 & 30.8 & 0.132 \\
\hline
\multicolumn{11}{|l|}{\textit{Gemma family}} \\
\hline
Gemma-2B        & \textbf{0.327} & \textbf{0.867} & \textbf{3.584} & \textbf{15.4} & \textbf{0.328} & 0.757 & 0.845 & 4.098 & 12.7 & 0.057 \\
Gemma-7B        & \textbf{0.053} & \textbf{0.973} & \textbf{3.758} & \textbf{2.43} & \textbf{0.254} & 0.200 & 0.895 & 4.397 & 1.79 & 0.108 \\
\hline
\multicolumn{11}{|l|}{\textit{LLaMA family}} \\
\hline
LLaMA-2-7B      & \textbf{0.007} & \textbf{0.996} & \textbf{3.813} & \textbf{6.33} & \textbf{0.183} & 0.032 & 0.975 & 4.543 & 5.36 & 0.048 \\
LLaMA-2-13B     & \textbf{0.055} & \textbf{0.967} & \textbf{1.635} & \textbf{6.90} & \textbf{0.275} & 0.221 & 0.868 & 2.106 & 5.16 & 0.044 \\
LLaMA-3-8B      & \textbf{0.045} & \textbf{0.973} & \textbf{3.755} & \textbf{1.95} & \textbf{0.263} & 0.197 & 0.893 & 4.580 & 1.57 & 0.072 \\
LLaMA-3.1-8B    & \textbf{0.038} & \textbf{0.977} & \textbf{3.714} & \textbf{2.41} & \textbf{0.265} & 0.168 & 0.907 & 4.543 & 2.06 & 0.064 \\
\hline
\end{tabular}
\end{table*}

\subsection{RQ1: Multi-Domain Validation and Classifier Discriminability}
\label{sec:rq1_multidomain}

\paragraph{Four-way domain comparison.}
To test whether the HP-vs-TOFU separation reflects a structural known/unknown difference rather than an artifact of one narrow contrast, we extend the comparison to Harry Potter (retain split) and Star Wars, each tested against the same TOFU baseline across all 9 models (fresh, direction-corrected extraction), with two-sided Wilcoxon signed-rank tests and BH/FDR correction applied jointly across all 24 (metric, pair) tests.

\begin{table}[ht]
\centering
\small
\setlength{\tabcolsep}{3pt}
\caption{
    Four-way domain comparison: Cohen's $d$ ($p$, BH-corrected) for each condition vs.\ TOFU, $n{=}9$ models.
    All 24 cells are BH-significant ($q\in[0.0063,0.0260]$).
}
\label{tab:fourway}
\begin{tabular}{|l|l|c|}
\hline
Metric & vs.\ TOFU & $d$ ($p$) \\
\hline\hline
\multirow{3}{*}{$\|\overline{\Delta\hat{v}}\|$ / Global\_Coh.}
  & HP-forget & $+1.58$ (.0039) \\
  & HP-retain & $+3.13$ (.0039) \\
  & Star Wars & $+1.69$ (.0039) \\
\hline
\multirow{3}{*}{Batch\_SecondMoment\_TopEig}
  & HP-forget & $+2.57$ (.0039) \\
  & HP-retain & $+6.20$ (.0039) \\
  & Star Wars & $+4.62$ (.0039) \\
\hline
\multirow{3}{*}{AUSS$_\text{L2}$ / Batch\_Cov\_Trace}
  & HP-forget & $-1.55$ (.0039) \\
  & HP-retain & $-3.35$ (.0039) \\
  & Star Wars & $-1.43$ (.0117) \\
\hline
\multirow{3}{*}{AUSS$_{\cos^2}$}
  & HP-forget & $-0.92$ (.0195) \\
  & HP-retain & $-1.59$ (.0039) \\
  & Star Wars & $-2.00$ (.0039) \\
\hline
\multirow{3}{*}{AUSS$_\text{Jac}$}
  & HP-forget & $-1.87$ (.0039) \\
  & HP-retain & $-2.09$ (.0039) \\
  & Star Wars & $-2.82$ (.0039) \\
\hline
\multirow{3}{*}{Batch\_Cov\_TopEig}
  & HP-forget & $-1.34$ (.0078) \\
  & HP-retain & $-2.00$ (.0039) \\
  & Star Wars & $-1.36$ (.0039) \\
\hline
\end{tabular}
\end{table}

Both additional ``known'' domains (HP-retain and Star Wars) separate from TOFU on every metric, in the same sign as the HP-forget contrast; HP-retain effect sizes are, if anything, larger (up to $d{=}6.20$) than the HP-forget contrast itself.

\paragraph{Classifier-based discriminability.}
The discriminability analysis in Table~\ref{tab:rq1_disc} reports $\Delta$/CI, which depends on the scale of the underlying metric. To confirm the separation is not an artifact of that scale, we build one sample per (model, domain) pair from the 8-dimensional standardized AUSS peak-layer feature vector and train Logistic Regression and a small MLP (8 hidden units) under leave-one-model-out cross-validation, so both domains of a held-out model are always scored by a classifier that never saw that model.

\begin{table}[ht]
\centering
\small
\caption{Classification performance on known (HP) vs.\ unknown (TOFU) domains using standardized 8-dimensional AUSS features, LOMO-CV, $n{=}9$ models (18 samples).}
\label{tab:classifier}
\begin{tabular}{|l|c|c|}
\hline
 & Logistic Regr. & MLP (8 hidden) \\
\hline\hline
Accuracy & 100\% & 94.4\% \\
AUC & 1.00 & 1.00 \\
Correct & 18/18 & 17/18 \\
Models fully correct & 9/9 & 8/9 \\
\hline
\end{tabular}
\end{table}

Both classifiers achieve near-perfect discriminability, with zero per-model misclassification for Logistic Regression, confirming the known/unknown separation is a robust structural feature of the representations rather than an artifact of the metrics' raw scale.

\subsection{RQ2: Full Geometry vs ASR Correlation Table}

See the full results at Tables~\ref{tab:rq2_main_llama2}--\ref{tab:rq2_main_qwen} and Table~\ref{tab:rq2_corr}.
\begin{table*}[t]
\centering
\small
\setlength{\tabcolsep}{4pt}
\caption{
    Spearman $\rho$ between all fragmentation metrics and attack-success rate across $n{=}44$--$45$ post-unlearning runs (excluding Base and WHP).
    ${}^{**}$/${}^{*}$/${}^{\dagger}$: BH-adjusted $p<0.01/0.05/0.10$ within each attack column (9 metrics).
    ${}^\star$: adaptation of Gram-matrix entropy to concept-differential vectors~\citep{skean_layer_2025}.
    \textbf{Bold}: survives BH at $p<0.05$.
    Only bold, starred cells ($q{<}0.05$) represent statistically robust findings; unmarked values are shown for completeness so the full 9-metric $\times$ 4-attack grid is auditable, and should not be read as evidence of a relationship.
    None of MIA's AUSS/AnonDiff correlations survive BH correction; see Section~\ref{sec:experiments}.
}

\label{tab:rq2_corr}
\begin{tabular}{|l||c|c|c|c|}
\hline
Metric & Steering & ICL & GCG & MIA \\
\hline\hline
\multicolumn{5}{|l|}{\textit{AUSS dispersion}} \\
\hline
AUSS$_\text{L2}$                    & $+0.00$ & $\mathbf{-0.51}^{**}$ & $-0.06$ & $+0.04$ \\
AUSS$_{\cos^2}$                     & $+0.06$ & $\mathbf{-0.50}^{**}$ & $-0.01$ & $+0.10$ \\
$\|\overline{\Delta\hat{v}}\|$           & $-0.00$ & $\mathbf{+0.51}^{**}$ & $+0.06$ & $-0.04$ \\
\hline
\multicolumn{5}{|l|}{\textit{AnonDiff}} \\
\hline
AnonDiff~$\overline{\cos}$          & $-0.24$ & $+0.27$ & $-0.09$ & $-0.15$ \\
AnonDiff~$S_2^{\star}$              & $\mathbf{+0.48}^{**}$ & $-0.22$ & $+0.22$ & $+0.20$ \\
AnonDiff~$\bar{L}_2$                & $+0.15$ & $+0.04$ & $-0.05$ & $-0.07$ \\
\hline
\multicolumn{5}{|l|}{\textit{Reg (raw-activation baseline)}} \\
\hline
Reg~$S_2^{\star}$                   & $+0.27$ & $+0.05$ & $\mathbf{+0.45}^{**}$ & $+0.29$ \\
Reg~$\overline{\cos}$               & $-0.31$ & $+0.01$ & $\mathbf{-0.51}^{**}$ & $-0.27$ \\
Reg~$\bar{L}_2$                     & $-0.07$ & $-0.09$ & $-0.24$ & $-0.16$ \\
\hline
\end{tabular}
\end{table*}

\subsection{Head-to-Head Against Xu et al.\ (2026)}
\label{sec:xu_comparison}

Full 44-test comparison (8 AUSS metrics + 3 Xu et al.\ metrics $\times$ 4 attacks) underlying the main-text summary in Section~\ref{sec:experiments}, BH-corrected globally across all 44 tests, $n{=}53$--$54$.
AUSS values come from checkpoints retrained specifically for this comparison (the originals were already deleted at the time this analysis was run), paired with the original checkpoints' attack scores; this table should be read as an independent second data point rather than checkpoint-consistent with Table~\ref{tab:rq2_corr}.

\begin{table}[ht]
\centering
\tiny
\setlength{\tabcolsep}{3pt}
\caption{Full Xu et al.\ head-to-head comparison, sorted by $q_\text{global}$. ${}^{**}/{}^{*}$: $q{<}0.01/0.05$.}
\label{tab:xu_comparison}
\begin{tabular}{|l|l|l|c|c|c|}
\hline
Family & Metric & Attack & $\rho$ & $p$ & $q_\text{global}$ \\
\hline\hline
AUSS & Centroid\_Norm & ICL & +0.551 & 0.00002 & 0.0001\textsuperscript{**} \\
AUSS & AUSS$_{\cos^2}$ & ICL & $-$0.572 & 0.00001 & 0.0001\textsuperscript{**} \\
AUSS & AUSS$_{L2}$ & ICL & $-$0.551 & 0.00002 & 0.0001\textsuperscript{**} \\
AUSS & Global\_Coherence & ICL & +0.551 & 0.00002 & 0.0001\textsuperscript{**} \\
AUSS & Batch\_Cov\_Trace & ICL & $-$0.551 & 0.00002 & 0.0001\textsuperscript{**} \\
AUSS & Batch\_SecMom\_TopEig & ICL & +0.557 & 0.00001 & 0.0001\textsuperscript{**} \\
AUSS & AUSS$_{Jac}$ & Steering & $-$0.452 & 0.00067 & 0.0042\textsuperscript{**} \\
Xu et al. & cka\_mean & Steering & +0.428 & 0.00137 & 0.0076\textsuperscript{**} \\
Xu et al. & pca\_sim\_mean & ICL & +0.381 & 0.00485 & 0.0212\textsuperscript{*} \\
AUSS & Batch\_Cov\_TopEig & ICL & $-$0.385 & 0.00440 & 0.0212\textsuperscript{*} \\
Xu et al. & cka\_mean & GCG & +0.374 & 0.00529 & 0.0212\textsuperscript{*} \\
Xu et al. & cka\_mean & MIA & +0.336 & 0.01378 & 0.0505 \\
AUSS & Batch\_Cov\_TopEig & GCG & $-$0.315 & 0.02020 & 0.0684 \\
Xu et al. & cka\_mean & ICL & +0.310 & 0.02409 & 0.0757 \\
AUSS & Batch\_Cov\_TopEig & Steering & $-$0.287 & 0.03693 & 0.1083 \\
Xu et al. & pca\_sim\_mean & MIA & +0.269 & 0.05126 & 0.1410 \\
Xu et al. & pca\_sim\_mean & GCG & +0.263 & 0.05497 & 0.1423 \\
AUSS & Centroid\_Norm & GCG & +0.218 & 0.11245 & 0.2262 \\
AUSS & Batch\_Cov\_Trace & GCG & $-$0.218 & 0.11245 & 0.2262 \\
AUSS & Global\_Coherence & GCG & +0.218 & 0.11245 & 0.2262 \\
AUSS & AUSS$_{Jac}$ & GCG & $-$0.218 & 0.11308 & 0.2262 \\
AUSS & AUSS$_{L2}$ & GCG & $-$0.218 & 0.11245 & 0.2262 \\
Xu et al. & pca\_shift\_mean & MIA & +0.213 & 0.12546 & 0.2300 \\
AUSS & Batch\_SecMom\_TopEig & GCG & +0.212 & 0.12336 & 0.2300 \\
Xu et al. & pca\_shift\_mean & GCG & $-$0.205 & 0.13778 & 0.2425 \\
AUSS & AUSS$_{\cos^2}$ & GCG & $-$0.189 & 0.17119 & 0.2897 \\
AUSS & Batch\_Cov\_Trace & Steering & $-$0.110 & 0.43311 & 0.5775 \\
AUSS & Global\_Coherence & Steering & +0.110 & 0.43311 & 0.5775 \\
Xu et al. & pca\_sim\_mean & Steering & $-$0.114 & 0.41497 & 0.5775 \\
AUSS & AUSS$_{L2}$ & Steering & $-$0.110 & 0.43311 & 0.5775 \\
Xu et al. & pca\_shift\_mean & ICL & +0.122 & 0.38399 & 0.5775 \\
AUSS & Centroid\_Norm & Steering & +0.110 & 0.43311 & 0.5775 \\
AUSS & Batch\_SecMom\_TopEig & Steering & +0.111 & 0.42803 & 0.5775 \\
AUSS & Batch\_SecMom\_TopEig & MIA & $-$0.079 & 0.57261 & 0.6460 \\
AUSS & Batch\_Cov\_Trace & MIA & +0.080 & 0.56753 & 0.6460 \\
AUSS & AUSS$_{\cos^2}$ & MIA & +0.085 & 0.54492 & 0.6460 \\
AUSS & Global\_Coherence & MIA & $-$0.080 & 0.56753 & 0.6460 \\
AUSS & AUSS$_{L2}$ & MIA & +0.080 & 0.56753 & 0.6460 \\
AUSS & Centroid\_Norm & MIA & $-$0.080 & 0.56753 & 0.6460 \\
AUSS & Batch\_Cov\_TopEig & MIA & +0.066 & 0.63949 & 0.7034 \\
AUSS & AUSS$_{\cos^2}$ & Steering & $-$0.054 & 0.69991 & 0.7511 \\
AUSS & AUSS$_{Jac}$ & ICL & $-$0.051 & 0.71931 & 0.7536 \\
AUSS & AUSS$_{Jac}$ & MIA & +0.032 & 0.81909 & 0.8381 \\
Xu et al. & pca\_shift\_mean & Steering & $-$0.011 & 0.93846 & 0.9385 \\
\hline
\end{tabular}
\end{table}

\subsection{Layerwise Geometry Analysis}
\label{sec:layer_analysis}

\begin{figure*}[t]
\centering
\includegraphics[width=\textwidth]{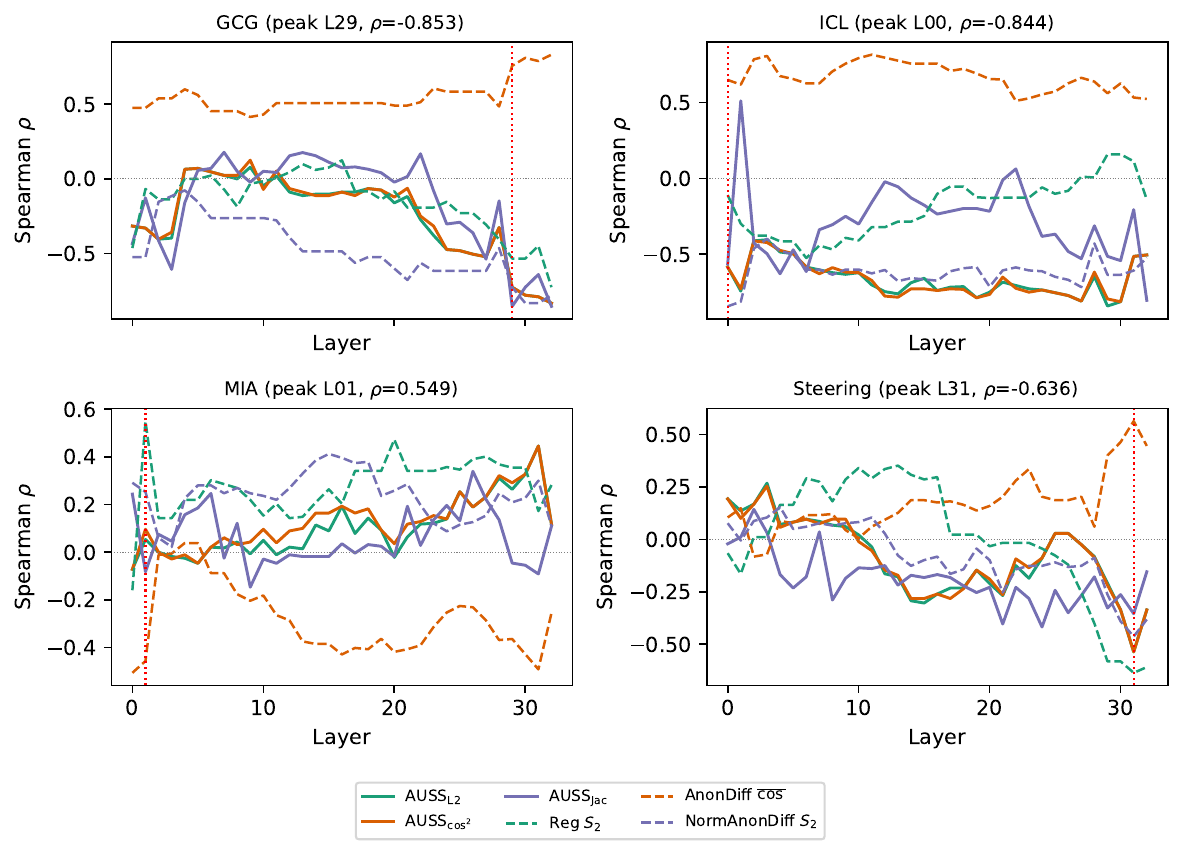}
\caption{
    \textbf{Spearman $\rho$ by layer for all four attacks, leader metrics.}
    Solid lines: AUSS family (AUSS$_\text{L2}$, AUSS$_{\cos^2}$, AUSS$_\text{Jac}$).
    Dashed lines: Entropy family (Reg~$S_2$, AnonDiff~$\overline{\cos}$, NormAnonDiff~$S_2$).
    Dotted red vertical line: attack-specific peak layer among the six plotted metrics (GCG→L29, ICL→L00 for this restricted set; see Table~\ref{tab:layer_best} for the full-metric-universe peaks used in the main text, GCG→L29, ICL→L28).
    GCG and ICL both concentrate their strongest signal late in the network; MIA and Steering show weaker, non-BH-significant layer-specific signal in this dataset.
    $n{=}10$--$15$ runs per attack (complete data).
}
\label{fig:layer_rho_all}
\end{figure*}

\paragraph{Methodology.}
Following established practice in mechanistic interpretability (e.g., causal tracing;~\citealt{meng_locating_2023}), we treat this layer-resolved sweep as an exploratory diagnostic map of \emph{where} different adversarial mechanisms exploit the geometry, rather than a multiplicity-corrected hypothesis test for a new global predictor; the corrected significance counts below quantify how much of the raw signal survives once treated as the latter.
We extend the correlation analysis of Section~\ref{sec:experiments} to every transformer layer.
For each run we extract hidden states at \emph{all} $L$ layers in a single forward pass with \texttt{output\_hidden\_states=True} (mean-pooled over the response tokens), giving vectors $\mathbf{h}^{(\ell)}_{\text{reg}}$ and $\mathbf{h}^{(\ell)}_{\text{anon}}$ at each layer $\ell \in \{0,\ldots,L{-}1\}$.
We compute Spearman~$\rho$ between each (metric, layer) pair and each attack-success score across all runs with complete data.
Two metric families are evaluated in parallel:
\begin{itemize}
    \item \textbf{AUSS metrics} (8 metrics, defined in Section~\ref{sec:method}): Centroid\_Norm, AUSS$_\text{L2}$, AUSS$_{\cos^2}$, AUSS$_\text{Jac}$, Global\_Coherence, Batch Covariance, etc.
    \item \textbf{Entropy / AnonDiff metrics} (14 metrics): Gram-matrix Rényi-$\alpha$ entropy of the regular ($\text{Reg}$), concept-differential ($\text{AnonDiff}$), and row-normalised concept-differential ($\text{NormAnonDiff}$) hidden-state matrices, plus mean cosine and mean $L_2$ statistics.
\end{itemize}
Figure~\ref{fig:layer_rho_all} shows the full $\rho$-vs.-layer sweep for all four attacks using the six leader metrics (three AUSS, three entropy).
The GCG panel reproduces and extends Figure~\ref{fig:layer_rho_gcg} from the main text.

\paragraph{Best layer-metric predictors per attack.}
Refer to Table~\ref{tab:layer_best}.

\begin{table*}[t]
\centering
\small
\setlength{\tabcolsep}{4pt}
\caption{
    Best (layer, metric) predictor per attack in the layer-resolved analysis.
    $n$: number of runs with complete data at that layer.
    AUSS and entropy families are shown separately; the overall best metric across both is marked $\star$.
    $q_\text{attack}$: BH correction within the attack's own column (8 AUSS + entropy tests); $q_\text{global}$: BH correction across all 4{,}224 (layer, metric, attack) tests at once.
    ${}^{*}q{<}0.05$.
}
\label{tab:layer_best}
\begin{tabular}{|l|l||c|c|c|c|c|}
\hline
Attack & Family & Best layer & Best metric & $\rho$ & $p$ & $q_\text{attack}$ / $q_\text{global}$ \\
\hline\hline
\multirow{2}{*}{GCG} & AUSS    & L29 & AUSS$_\text{Jac}$               & $-0.853$ & $0.0008$ & $0.057$ / $0.046^{*}$ \\
                     & Entropy & L30 & NormAnonDiff~$S_2$ $\star$      & $-0.832$ & $0.0015$ & $0.057$ / $0.046^{*}$ \\
\hline
\multirow{2}{*}{ICL} & AUSS    & L28 & Batch\_Cov\_TopEig $\star$      & $-0.886$ & ${<}0.0001$ & $0.006^{*}$ / $0.009^{*}$ \\
                     & Entropy & L00 & AnonDiff~$S_2$                  & $-0.861$ & $0.0002$ & $0.011^{*}$ / $0.037^{*}$ \\
\hline
\multirow{2}{*}{MIA} & AUSS    & L31 & Centroid\_Norm                  & $-0.446$ & $0.170$ & $0.999$ / $0.581$ \\
                     & Entropy & L32 & Reg~$\bar{L}_2$ $\star$         & $+0.627$ & $0.039$ & $0.999$ / $0.254$ \\
\hline
\multirow{2}{*}{Steering} & AUSS    & L31 & $\|\overline{\Delta\hat{v}}\|$   & $+0.536$ & $0.089$ & $0.987$ / $0.401$ \\
                          & Entropy & L31 & Reg~$\overline{\cos}$ $\star$ & $+0.645$ & $0.032$ & $0.987$ / $0.224$ \\
\hline
\end{tabular}
\end{table*}

\paragraph{Key findings.}
\begin{enumerate}
    \item \textbf{GCG and ICL are the only two attacks whose layer-specific signal survives correction.} Both peaks fall late in the network and one layer apart (GCG at L29, ICL at L28); GCG survives global correction ($q{=}0.046$) but not the stricter per-attack-column correction ($q{=}0.057$), while ICL survives both. See the main-text Discussion for the mechanistic reading.

    \item \textbf{MIA and Steering do not survive correction at either scope.} Their best (layer, metric) cells are only nominally significant at the raw $p{<}0.05$ level ($q_\text{global}{\geq}0.22$ for both); we report them descriptively rather than as reliable predictors.

    \item \textbf{The aggregate and layer-resolved analyses find GCG signal via different metric families.} The aggregate correlations in Table~\ref{tab:rq2_corr} find GCG significant via the Reg (Gram-entropy) family (Reg~$\overline{\cos}$, $\rho{=}{-}0.51$); the layer-resolved sweep instead finds its strongest GCG signal via AUSS$_\text{Jac}$ at L29 ($\rho{=}{-}0.85$). Both are real but reflect different aspects of the geometry: the peak-layer AUSS metric (argmax over AUSS$_\text{L2}$) does not coincide with GCG's own predictive layer.

    \item \textbf{Positive $\rho$ for entropy metrics on MIA and Steering.} Where entropy-family metrics show their strongest (non-significant) signal, higher entropy correlates with \emph{higher} attack success, directionally opposite to the AUSS findings on GCG/ICL, reflecting that entropy measures spread of the concept-differential subspace while coherence measures its alignment.
\end{enumerate}

\subsection{Ranking Consistency of the Aggregate Correlations}
\label{sec:loo_stability}

To test whether the aggregate correlations in Table~\ref{tab:rq2_corr} are driven by one influential run, we compute leave-one-run-out (LOO) Spearman $\rho$ for each of the 8 AUSS metrics against each of the 4 attacks, restricted to verified unlearning runs ($n{=}26$--$27$).

\begin{table}[ht]
\centering
\small
\caption{Leave-one-out sign-flip stability, 8 AUSS metrics per attack, $n{=}26$--$27$ verified runs.}
\label{tab:loo_stability}
\begin{tabular}{|l|cc|}
\hline
Attack & \begin{tabular}[c]{@{}c@{}}Metrics with 0\\ LOO sign flips\end{tabular} & \begin{tabular}[c]{@{}c@{}}BH-significant\\ ($q{<}0.05$)\end{tabular} \\
\hline\hline
ICL & 8/8 & 7/8 \\
Steering & 8/8 & 0/8 \\
GCG & 7/8 & 0/8 \\
MIA & 2/8 & 0/8 \\
\hline
\end{tabular}
\end{table}

Only ICL is both leave-one-out stable and statistically significant after BH correction (8/8 metrics stable, 7/8 significant). Steering and GCG are largely stable (8/8 and 7/8) but not significant; MIA is neither reliably stable nor significant. We therefore present the correlation findings within the context of the ICL attack, rather than as a generalized cross-attack ranking claim.

\subsection{RQ2: Full Per-Run Results}
\label{sec:rq2_full_runs}

\begin{table*}[t]
\centering
\small
\setlength{\tabcolsep}{4pt}
\caption{
    \textbf{RQ2: Full post-unlearning results, LLaMA-2-7B-chat}.
    ROUGE-L$\downarrow$: forget-set ROUGE-L after unlearning.
    $\checkmark$: verified (ROUGE-L~$<0.10$ \& retain~$\geq 0.80\times$ base).
    $\|\overline{\Delta\hat{v}}\|{\uparrow}$: centroid norm; AUSS$_\text{L2}{\downarrow}$, AUSS$_{\cos^2}{\downarrow}$: dispersion (per-method peak layer $\ell^*$).
    Steer-AUC$\downarrow$: steering AUC; Steer-WF$\downarrow$: mean steered word-frequency score.
    ICL~$\Delta{\uparrow}$: ICL accuracy minus baseline.
    MIA$\downarrow$: membership inference AUC; GCG$\downarrow$: GCG ASR.
    Blank cells: checkpoint unavailable.
}
\label{tab:rq2_main_llama2}
\begin{tabular}{|ll|c||c||ccc||ccccc|}
\hline
 & & & Forget & \multicolumn{3}{c||}{Geometry $\ell^*$} & \multicolumn{5}{c|}{Attack Success $\downarrow$} \\
\cline{4-12}
Method & Cpt & $\checkmark$ & ROUGE-L$\downarrow$ & $\|\overline{\Delta\hat{v}}\|{\uparrow}$ & AUSS$_\text{L2}{\downarrow}$ & AUSS$_{\cos^2}{\downarrow}$ & Steer-AUC & Steer-WF & ICL~$\Delta$ & MIA & GCG \\
\hline\hline
 Base & HP &  & 0.131 & 0.878 & 0.507 & 0.439 & 0.428 & 0.265 & +0.000 & 0.827 & 0.000 \\
         & SW &  & 0.152 & 0.876 & 0.516 & 0.447 & 0.449 & 0.296 & +0.120 & 0.955 & 0.000 \\
         & WS &  & 0.090 & 0.911 & 0.379 & 0.341 & 0.503 & 0.274 & +0.040 & 0.837 & 0.000 \\
  \hline
 GA & HP & \checkmark & 0.000 & 0.823 & 0.717 & 0.553 & 0.500 & 0.000 & +0.000 & 0.678 & 0.000 \\
         & SW & \checkmark & 0.000 & 0.875 & 0.521 & 0.442 & 0.500 & 0.000 & +0.000 & 0.695 & 0.000 \\
         & WS & \checkmark & 0.000 & 0.855 & 0.598 & 0.475 & 0.020 & 0.000 & +0.000 & 0.615 & 0.000 \\
  \hline
 DPO & HP & \checkmark & 0.082 & 0.868 & 0.549 & 0.469 & 0.694 & 0.283 & $-$0.040 & 0.873 & 0.800 \\
         & SW & \checkmark & 0.098 & 0.866 & 0.554 & 0.474 & 0.673 & 0.281 & +0.040 & 0.986 & 0.800 \\
         & WS & \checkmark & 0.092 & 0.913 & 0.371 & 0.335 & 0.658 & 0.287 & +0.080 & 0.920 & 1.000 \\
  \hline
 NPO & HP &  & 0.127 & 0.860 & 0.580 & 0.492 & 0.756 & 0.286 & +0.000 & 0.805 & 0.600 \\
         & SW &  & 0.108 & 0.877 & 0.513 & 0.446 & 0.668 & 0.281 & +0.040 & 0.937 & 1.000 \\
         & WS & \checkmark & 0.092 & 0.922 & 0.333 & 0.303 & 0.795 & 0.295 & +0.040 & 0.748 & 1.000 \\
  \hline
 NPO+KL & HP & \checkmark & 0.098 & 0.854 & 0.603 & 0.507 & 0.769 & 0.259 & $-$0.040 & 0.449 & 0.600 \\
         & SW &  & 0.105 & 0.867 & 0.550 & 0.473 & 0.746 & 0.272 & +0.000 & 0.281 & 0.600 \\
         & WS & \checkmark & 0.091 & 0.917 & 0.354 & 0.320 & 0.823 & 0.286 & +0.040 & 0.162 & 1.000 \\
  \hline
 RMU & HP &  & 0.131 & 0.878 & 0.509 & 0.441 & 0.710 & 0.264 & +0.000 & 0.806 & 1.000 \\
         & SW &  & 0.143 & 0.875 & 0.519 & 0.449 & 0.694 & 0.300 & +0.040 & 0.951 & 1.000 \\
         & WS & \checkmark & 0.100 & 0.908 & 0.388 & 0.348 & 0.754 & 0.267 & +0.080 & 0.787 & 1.000 \\
  \hline
 WHP & HP &  & 0.255 & 0.858 & 0.585 & 0.494 & 0.756 & 0.248 & $-$0.040 & 0.442 & 0.000 \\
         & SW &  & 0.175 & 0.856 & 0.595 & 0.505 & 0.708 & 0.269 & +0.000 & 0.276 & 0.000 \\
         & WS &  & 0.135 & 0.917 & 0.353 & 0.319 & 0.826 & 0.281 & +0.040 & 0.181 & 0.000 \\
  \hline
\end{tabular}
\end{table*}

\begin{table*}[t]
\centering
\small
\setlength{\tabcolsep}{4pt}
\caption{
    \textbf{RQ2: Full post-unlearning results, Meta-LLaMA-3-8B-Instruct}.
    ROUGE-L$\downarrow$: forget-set ROUGE-L after unlearning.
    $\checkmark$: verified (ROUGE-L~$<0.10$ \& retain~$\geq 0.80\times$ base).
    $\|\overline{\Delta\hat{v}}\|{\uparrow}$: centroid norm; AUSS$_\text{L2}{\downarrow}$, AUSS$_{\cos^2}{\downarrow}$: dispersion (per-method peak layer $\ell^*$).
    Steer-AUC$\downarrow$: steering AUC; Steer-WF$\downarrow$: mean steered word-frequency score.
    ICL~$\Delta{\uparrow}$: ICL accuracy minus baseline.
    MIA$\downarrow$: membership inference AUC; GCG$\downarrow$: GCG ASR.
    Blank cells: checkpoint unavailable.
}
\label{tab:rq2_main_llama3}
\begin{tabular}{|ll|c||c||ccc||ccccc|}
\hline
 & & & Forget & \multicolumn{3}{c||}{Geometry $\ell^*$} & \multicolumn{5}{c|}{Attack Success $\downarrow$} \\
\cline{4-12}
Method & Cpt & $\checkmark$ & ROUGE-L$\downarrow$ & $\|\overline{\Delta\hat{v}}\|{\uparrow}$ & AUSS$_\text{L2}{\downarrow}$ & AUSS$_{\cos^2}{\downarrow}$ & Steer-AUC & Steer-WF & ICL~$\Delta$ & MIA & GCG \\
\hline\hline
 Base & HP &  & 0.062 & 0.856 & 0.594 & 0.493 & 0.423 & 0.437 & +0.000 & 0.915 & 0.000 \\
         & SW &  & 0.113 & 0.873 & 0.529 & 0.455 & 0.424 & 0.379 & +0.040 & 0.987 & 0.000 \\
         & WS &  & 0.068 & 0.901 & 0.416 & 0.370 & 0.425 & 0.347 & +0.040 & 0.952 & 0.000 \\
  \hline
 GA & HP & \checkmark & 0.000 & 0.872 & 0.534 & 0.439 & 0.500 & 0.000 & +0.000 & 0.700 & 0.000 \\
         & SW & \checkmark & 0.000 & 0.801 & 0.797 & 0.591 & 0.500 & 0.000 & +0.000 & 0.102 & 0.000 \\
         & WS & \checkmark & 0.000 & 0.814 & 0.750 & 0.436 & 0.500 & 0.000 & +0.000 & 0.250 & 0.000 \\
  \hline
 DPO & HP & \checkmark & 0.000 & 0.867 & 0.551 & 0.466 & 0.720 & 0.076 & +0.040 & 1.000 & 0.400 \\
         & SW & \checkmark & 0.017 & 0.842 & 0.646 & 0.519 & 0.620 & 0.017 & +0.000 & 0.998 & 0.800 \\
         & WS & \checkmark & 0.000 & 0.923 & 0.328 & 0.298 & 0.700 & 0.037 & +0.000 & 1.000 & 0.000 \\
  \hline
 NPO & HP &  & 0.112 & 0.863 & 0.566 & 0.482 & 0.930 & 0.493 & +0.000 & 0.817 & 1.000 \\
         & SW &  & 0.167 & 0.871 & 0.535 & 0.462 & 0.838 & 0.404 & $-$0.040 & 0.900 & 1.000 \\
         & WS & \checkmark & 0.096 & 0.923 & 0.328 & 0.298 & 0.779 & 0.356 & +0.080 & 0.834 & 1.000 \\
  \hline
 NPO+KL & HP &  & 0.178 & 0.867 & 0.552 & 0.472 & 0.945 & 0.488 & $-$0.120 & 0.389 & 1.000 \\
         & SW &  & 0.208 & 0.847 & 0.627 & 0.525 & 0.882 & 0.393 & $-$0.120 & 0.249 & 1.000 \\
         & WS &  & 0.121 & 0.921 & 0.335 & 0.305 & 0.822 & 0.313 & +0.080 & 0.157 & 1.000 \\
  \hline
 RMU & HP & \checkmark & 0.063 & 0.856 & 0.595 & 0.493 & 0.896 & 0.444 & $-$0.040 & 0.862 & 1.000 \\
         & SW &  & 0.123 & 0.871 & 0.538 & 0.460 & 0.854 & 0.394 & +0.040 & 0.925 & 1.000 \\
         & WS & \checkmark & 0.072 & 0.902 & 0.416 & 0.370 & 0.766 & 0.350 & +0.040 & 0.890 & 1.000 \\
  \hline
 WHP & HP &  & 0.174 & 0.859 & 0.583 & 0.493 & 0.922 & 0.483 & $-$0.080 & 0.415 & 0.000 \\
         & SW &  & 0.208 & 0.860 & 0.579 & 0.494 & 0.846 & 0.382 & $-$0.120 & 0.283 & 0.000 \\
         & WS &  & 0.103 & 0.926 & 0.316 & 0.288 & 0.850 & 0.334 & +0.000 & 0.178 & 0.000 \\
  \hline
\end{tabular}
\end{table*}

\begin{table*}[t]
\centering
\small
\setlength{\tabcolsep}{4pt}
\caption{
    \textbf{RQ2: Full post-unlearning results, Qwen2.5-7B-Instruct}.
    ROUGE-L$\downarrow$: forget-set ROUGE-L after unlearning.
    $\checkmark$: verified (ROUGE-L~$<0.10$ \& retain~$\geq 0.80\times$ base).
    $\|\overline{\Delta\hat{v}}\|{\uparrow}$: centroid norm; AUSS$_\text{L2}{\downarrow}$, AUSS$_{\cos^2}{\downarrow}$: dispersion (per-method peak layer $\ell^*$).
    Steer-AUC$\downarrow$: steering AUC; Steer-WF$\downarrow$: mean steered word-frequency score.
    ICL~$\Delta{\uparrow}$: ICL accuracy minus baseline.
    MIA$\downarrow$: membership inference AUC; GCG$\downarrow$: GCG ASR.
    Blank cells: checkpoint unavailable.
}
\label{tab:rq2_main_qwen}
\begin{tabular}{|ll|c||c||ccc||ccccc|}
\hline
 & & & Forget & \multicolumn{3}{c||}{Geometry $\ell^*$} & \multicolumn{5}{c|}{Attack Success $\downarrow$} \\
\cline{4-12}
Method & Cpt & $\checkmark$ & ROUGE-L$\downarrow$ & $\|\overline{\Delta\hat{v}}\|{\uparrow}$ & AUSS$_\text{L2}{\downarrow}$ & AUSS$_{\cos^2}{\downarrow}$ & Steer-AUC & Steer-WF & ICL~$\Delta$ & MIA & GCG \\
\hline\hline
 Base & HP &  & 0.106 & 0.872 & 0.534 & 0.439 & 0.433 & 0.373 & +0.040 & 0.982 & 0.000 \\
         & SW &  & 0.098 & 0.866 & 0.557 & 0.473 & 0.482 & 0.261 & +0.080 & 0.982 & 0.000 \\
         & WS &  & 0.080 & 0.907 & 0.395 & 0.352 & 0.495 & 0.373 & +0.000 & 0.938 & 0.000 \\
  \hline
 GA & HP & \checkmark & 0.000 & 0.936 & 0.275 & 0.253 & 0.500 & 0.000 & +0.000 & 0.119 & 0.000 \\
         & SW & \checkmark & 0.000 & 0.901 & 0.417 & 0.370 & 0.500 & 0.000 & +0.000 & 0.461 & 0.000 \\
         & WS & \checkmark & 0.000 & 0.934 & 0.284 & 0.255 & 0.500 & 0.000 & +0.000 & 0.185 & 0.200 \\
  \hline
 DPO & HP & \checkmark & 0.004 & 0.884 & 0.484 & 0.422 & 0.730 & 0.194 & +0.000 & 0.861 & 0.000 \\
         & SW &  & 0.117 & 0.834 & 0.676 & 0.548 & 0.677 & 0.343 & +0.000 & 0.965 & 1.000 \\
         & WS & \checkmark & 0.020 & 0.895 & 0.441 & 0.389 & 0.911 & 0.238 & +0.040 & 0.871 & 0.000 \\
  \hline
 NPO & HP &  & 0.198 & 0.863 & 0.569 & 0.478 & 0.744 & 0.374 & +0.000 & 0.805 & 1.000 \\
         & SW &  & 0.189 & 0.846 & 0.632 & 0.524 & 0.782 & 0.324 & +0.000 & 0.871 & 0.000 \\
         & WS &  & 0.105 & 0.913 & 0.370 & 0.331 & 0.706 & 0.433 & +0.040 & 0.770 & 1.000 \\
  \hline
 NPO+KL & HP &  & 0.349 & 0.865 & 0.559 & 0.474 & 0.934 & 0.458 & +0.000 & 0.482 & 0.000 \\
         & SW &  & 0.167 & 0.838 & 0.662 & 0.542 & 0.822 & 0.210 & +0.000 & 0.246 & 0.800 \\
         & WS &  & 0.145 & 0.908 & 0.390 & 0.348 & 0.874 & 0.449 & +0.000 & 0.218 & 0.000 \\
  \hline
 RMU & HP & \multicolumn{10}{c|}{---} \\
         & SW & \checkmark & 0.093 & 0.783 & 0.860 & 0.354 & 0.784 & 0.263 & +0.000 & 0.854 & 1.000 \\
         & WS & \checkmark & 0.082 & 0.875 & 0.522 & 0.439 & 0.634 & 0.364 & +0.040 & 0.785 & 0.800 \\
  \hline
 WHP & HP &  & 0.258 & 0.864 & 0.564 & 0.477 & 0.943 & 0.310 & +0.000 & 0.495 & 0.000 \\
         & SW &  & 0.172 & 0.843 & 0.644 & 0.533 & 0.845 & 0.205 & +0.000 & 0.218 & 0.000 \\
         & WS &  & 0.207 & 0.900 & 0.423 & 0.373 & 0.881 & 0.405 & +0.040 & 0.245 & 0.000 \\
  \hline
\end{tabular}
\end{table*}